\pdfoutput=1
\documentclass[11pt]{article}

\usepackage{acl}

\usepackage{times}
\usepackage{latexsym}

\usepackage[T1]{fontenc}
\usepackage[utf8]{inputenc}

\usepackage{microtype}

\usepackage{inconsolata}

\usepackage{graphicx}

\usepackage{pifont}

\usepackage{color, colortbl}
\definecolor{Gray}{gray}{0.93}
\definecolor{Orange}{rgb}{1,0.5,0}
\definecolor{DGray}{gray}{0.83}
\definecolor{LightCyan}{rgb}{0.88,1,1}

\definecolor{natural}{HTML}{648FFF}
\definecolor{specialized}{HTML}{DC267F}
\definecolor{fgvc}{HTML}{362682}
\definecolor{others}{HTML}{e6c595}
\definecolor{all}{HTML}{FE6100}

\definecolor{lightorange}{HTML}{fc8e62}
\definecolor{lightgray}{gray}{0.6}

\usepackage{color, colortbl}
\usepackage{soul}
\definecolor{Gray}{gray}{0.93}
\definecolor{Orange}{rgb}{1,0.5,0}
\definecolor{DGray}{gray}{0.83}
\definecolor{LightCyan}{rgb}{0.88,1,1}

\definecolor{WarnREd}{rgb}{1,0.4,0.4}
\definecolor{WarnOrange}{rgb}{1,0.682,0.502}
\definecolor{WarnPink}{rgb}{0.9176, 0.7215, 0.7215}
\definecolor{GoodGreen}{rgb}{0.5019, 0.9215, 0.6039}

\newcommand{\congrat}[1]{\sethlcolor{GoodGreen}\hl{#1}}

\newcommand{\YH}[1]{\textcolor{orange}{YH: [#1]}}

\newcommand{\npo}{\textsc{NPO}}
\newcommand{\rmu}{\textsc{RMU}}
\newcommand{\ga}{\textsc{GA}}
\newcommand{\gdiff}{\textsc{GDiff}}
\newcommand{\ours}{\textsc{SEUF}}

\newcommand{\wmdp}{\texttt{WMDP}}

\newcommand{\rwku}{\texttt{RWKU}}

\usepackage{wrapfig}
\usepackage{verbatim}
\usepackage{pifont}
\usepackage{color, colortbl}

\usepackage{blindtext}
\usepackage{lipsum}

\usepackage{multirow}
\usepackage{graphicx}
\usepackage{listings}

\usepackage{bbm}

\usepackage[most]{tcolorbox}

\usepackage{tikz}
\usepackage{tabularx}
\usepackage{amsmath,amsfonts,bm}
\usepackage{amsmath,amsfonts,bm}

\def\eqref#1{(\ref{#1})}
\def\1{\bm{1}}

\DeclareMathAlphabet{\mathsfit}{\encodingdefault}{\sfdefault}{m}{sl}
\SetMathAlphabet{\mathsfit}{bold}{\encodingdefault}{\sfdefault}{bx}{n}

\newcommand{\btheta}{\boldsymbol{\theta}}

\newcommand{\bx}{\mathbf{x}}

\usepackage{hyperref}
\usepackage{url}
\usepackage{graphicx}
\usepackage{subfigure}
\usepackage{booktabs}
\usepackage{caption}
\usepackage{algorithm}
\usepackage{algorithmic}

\definecolor{ceruleanblue}{rgb}{0.16, 0.32, 0.75}

\hypersetup{
colorlinks=true,
citecolor=ceruleanblue,
linkcolor=ceruleanblue,
urlcolor=black}

\everydisplay{\small} %%% equation in small font

\title{\ours: Is Unlearning One Expert Enough for Mixture-of-Experts LLMs?}

\vspace{-0.2in}\author{Haomin Zhuang$^{1*}$, Yihua Zhang$^2$\thanks{Equal contribution}, Kehan Guo$^1$, \\
\textbf{Jinghan Jia}$^2$, \textbf{Gaowen Liu}$^3$, \textbf{Sijia Liu}$^2$, \textbf{Xiangliang Zhang}$^1$ \\
\small$^1$University of Notre Dame
\small$^2$Michigan State University
\small$^3$Cisco Research\\
\small \texttt{\{hzhuang2,xzhang33\}@nd.edu}
\vspace{-6in}}

\begin{document}

\maketitle
\vspace{-2in}
\begin{abstract}
\vspace{-0.05in}
Recent advancements in LLMs 
unlearning have shown remarkable success in removing unwanted data-model influences while preserving the model's utility for legitimate knowledge. Despite these strides, sparse Mixture-of-Experts (MoE) LLMs--a key subset of the LLM family--have remain unexplored 
in the context of unlearning. As MoE LLMs are celebrated for their exceptional performance,  
we ask: \emph{How can unlearning be performed effectively and efficiently on MoE LLMs?} 
Our pilot study shows that the dynamic routing nature of MoE LLMs introduces unique challenges, leading to excessive forgetting,  uncontrolled knowledge erasure and 
substantial utility drops when existing unlearning methods are applied.
To address this, we propose a novel Selected-Expert Unlearning Framework ({\ours}). %, for MoE LLMs. 
Through expert attribution, unlearning is concentrated on the most actively engaged experts for the specified knowledge. Concurrently, an anchor loss is applied to the router to stabilize the active state of this targeted expert, ensuring focused and controlled unlearning. 
{\ours} is  compatible with various standard unlearning algorithms. Extensive experiments demonstrate that {\ours} enhances both forget quality up to $5\%$ and model utility by $35\%$ on MoE LLMs across various benchmarks and  LLM architectures (compared to standard unlearning algorithms), while only unlearning $0.06\%$ of the model parameters. 
\end{abstract}
\section{Introduction}
\label{sec: intro}
\vspace{-0.05in}
Despite the extraordinary ability in generating human-like content \citep{touvron2023llama}, the rapid development of large language models (LLMs) have raised a series of ethical and security concerns, such as pretraining on copyrighted data \citep{sun2024trustllm}, bias perpetuation \citep{motoki2023more}, the generation of toxic, biased, or illegal contents \citep{wen2023unveiling}, and facilitating making cyberattacks and bio-weapons \citep{li2024wmdp}. As a solution, the problem of Machine Unlearning (MU) arises (also referred to LLM unlearning) \citep{liu2024rethinking}, aiming to scrub the influence of the undesired training data and removing their corresponding generation abilities, while preserving the influence of other remaining valid data \citep{jia2024wagle, shi2024muse, jia2024soul}.

% LLMs are often trained on massive datasets that may unintentionally include private content and harmful information~\citep{liu2024rethinking}. Recent research have further underscored the need to remove proprietary data as well as malicious actions from models~\citep{li2024wmdp,jin2024rwku}. This has led to a growing interest in developing Machine Unlearning (MU) algorithms, which enable models to function as if specific data were never part of their training. Achieving exact unlearning requires retraining the model entirely without the unwanted data, a process that is resource-intensive and impractical for frequent use. To address this, multiple unlearning methods have been proposed~\citep{liu2022continual,zhang2024negative}.

While LLM unlearning has recently become a major research thrust, past efforts have only focused on effective unlearning methods for conventional LLMs. In contrast, sparse Mixture-of-Experts LLM (MoE LLM) \citep{jiang2024mixtral,grok1, databricks_dbrx_2024,abdin2024phi,liu2024deepseek}, designed to reduce computational burdens during inference, remained unexplored in this context. As a key member of the LLM family, MoE LLMs offer substantial scalability without a corresponding increase in computational costs \citep{jiang2024mixtral, qwen_moe, dai2024deepseekmoe}. Thanks to their dynamic routing mechanism, MoE LLMs direct inference through different model components, known as `experts'. However, it remains unclear how LLM unlearning interacts with the sparse MoE architecture and whether unlearning for MoE LLMs presents unique challenges. This leads us to ask:

% However, existing unlearning algorithms are primarily designed for dense LLMs, with minimal research on MU for Mixture of Experts (MoE) architecture LLMs. MoE LLMs offer scalability by activating only a subset of model parameters during inference, which enhances performance while reducing computational costs. MoE has sustained strong growth~\citep{cai2024survey}, especially evident in 2024, with the release of several large-scale industrial LLMs, including Grok-1~\citep{grok1}, DBRX~\citep{databricks_dbrx_2024}, Phi-3.5-MoE~\citep{abdin2024phi} and DeepSeek-v2~\citep{liu2024deepseek}. Given the growing number of MoE models, research on MU in MoE architectures is increasingly critical to address their security concerns.

% \vspace*{-3mm}
\begin{tcolorbox}[before skip=2mm, after skip=0.0cm, boxsep=0.0cm, middle=0.0cm, top=0.1cm, bottom=0.1cm]
    \textit{\textbf{(Q)} Can we develop a principled MU method for MoE LLMs that 
    ensures high forgetting effectiveness, while maintaining model utility and efficiency?}
\end{tcolorbox}
\vspace*{0.2cm}

\begin{figure*}[t]
\vspace*{-1em}
    \centering
    \begin{tabular}{cc}
         \hspace*{-3mm}\includegraphics[width=0.38\linewidth]{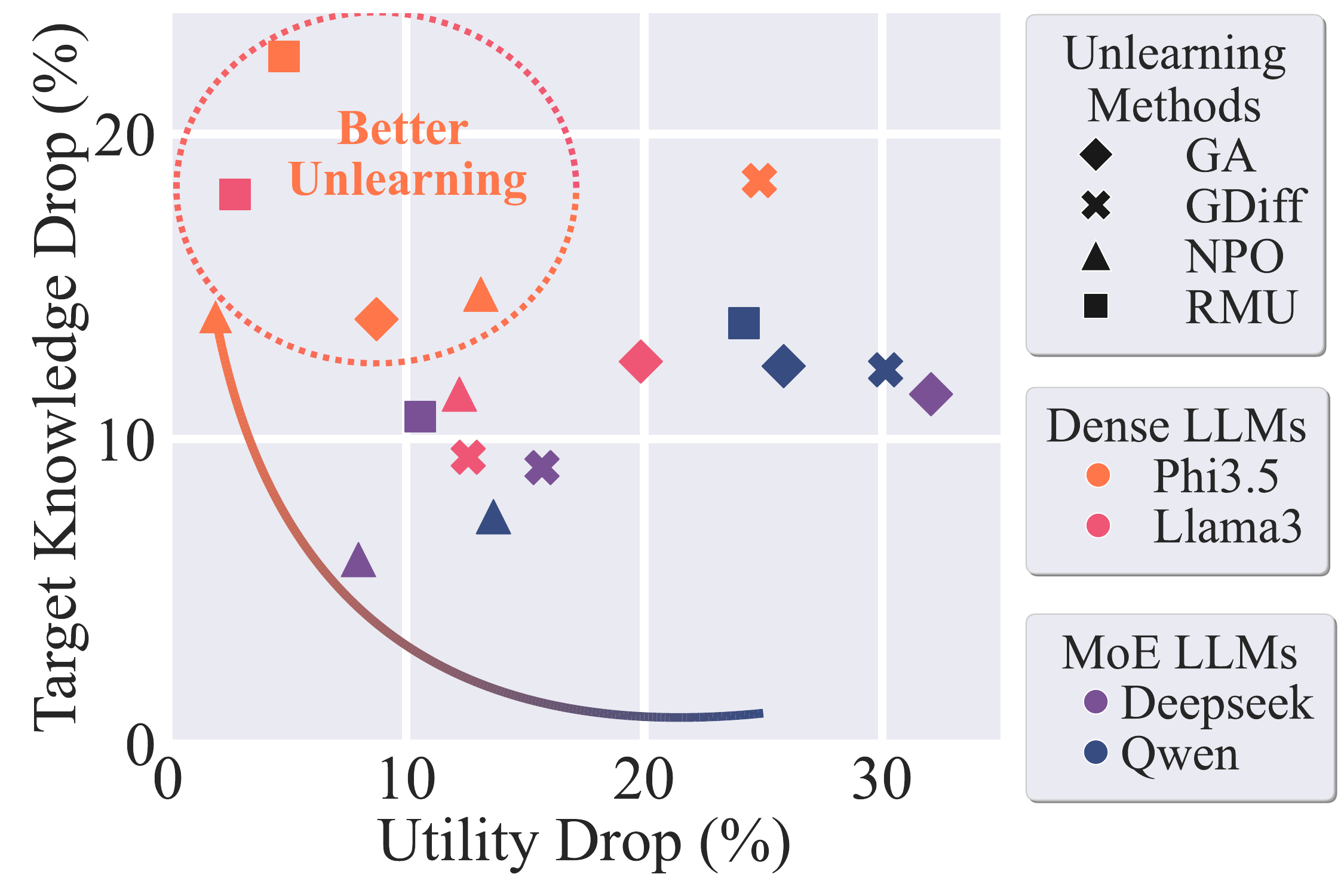} 
         & \hspace*{-1mm}\includegraphics[width=0.62\linewidth]{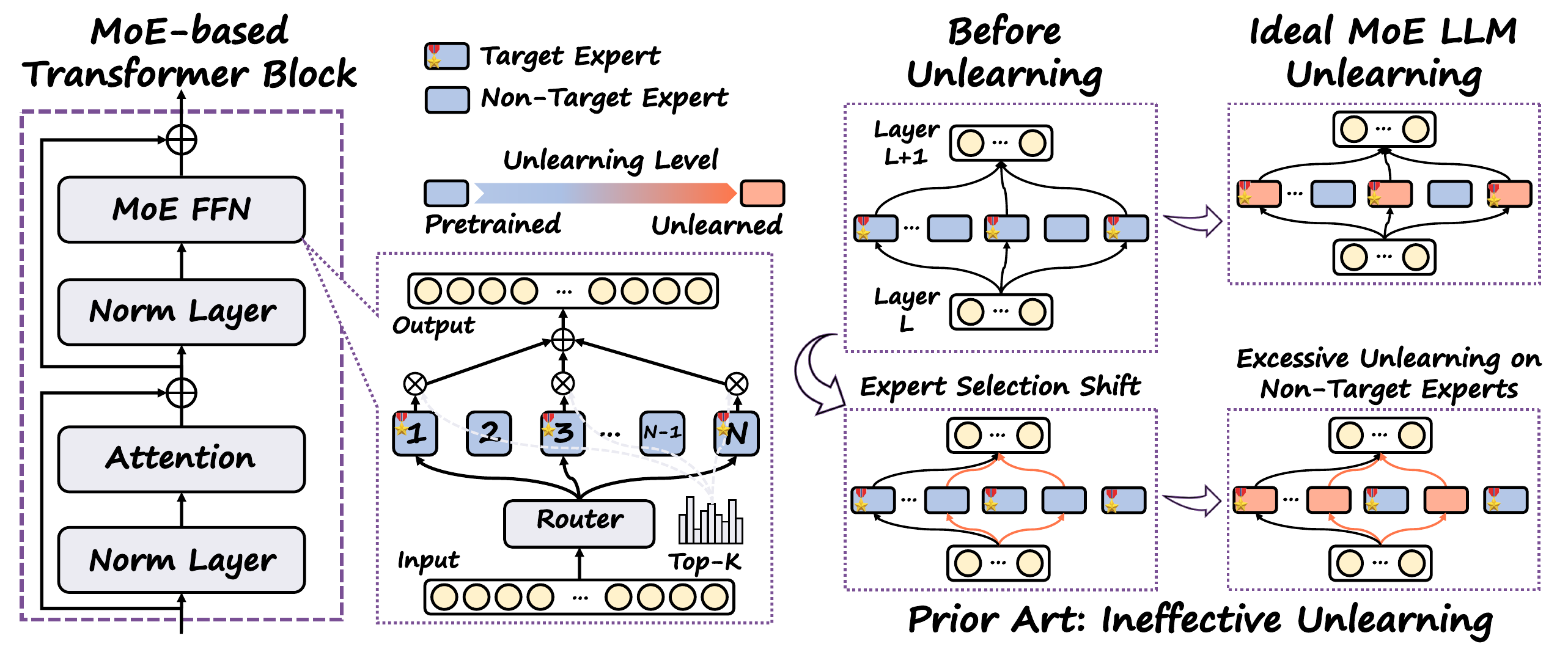}\vspace*{-2mm}\\
         \hspace*{-3mm}\footnotesize{(a)} & \hspace*{-1mm} \footnotesize{(b)}
    \end{tabular}
    \vspace*{-1.2em}
    \caption{\footnotesize Overview of the key findings in this paper. (a) Illustration of the ineffectiveness of existing unlearning methods on MoE LLMs. Four unlearning algorithms—{\ga} \citep{eldan2023whos}, {\gdiff} \citep{maini2024tofu}, {\npo} \citep{zhang2024negative}, and {\rmu} \citep{li2024wmdp}—were applied to two MoE LLMs (DeepSeek-v2-Lite \citep{liu2024deepseek} and Qwen1.5-MoE \citep{qwen_moe}) and two dense LLMs (Phi3.5 \citep{abdin2024phi} and LLaMA3-8B \citep{dubey2024llama}) using the {\wmdp} benchmark \cite{li2024wmdp}. The drop in target knowledge (accuracy drop on the forget test set, higher is better) and the drop in model utility (accuracy drop on MMLU \cite{hendrycks2023overview}, lower is better) are plotted. Better-unlearned models should appear  in the top left corner, but unlearning on MoE LLMs was less effective compared to non-MoE modles. (b) Illustration of ideal versus ineffective MoE LLM unlearning. Target experts—those most frequently activated given the forget set—are identified for unlearning. However, existing unlearning algorithms tend to cause substantial expert selection shifts, leading to excessive and unnecessary unlearning of non-target experts, which significantly impairs model utility.}
    \label{fig: teaser}
    \vspace*{-1.0em}
\end{figure*}

To the best of our knowledge, the problem (\textbf{Q}) remains unexplored in the current literature. Our investigation begins with a pilot study that applies existing unlearning methods to MoE LLMs. Preliminary results indicate that a simple application of these methods can lead to a substantial drop in model utility and even model collapse. This phenomenon is illustrated in \textbf{Fig.\,\ref{fig: teaser}}(a), which depicts the performance of the unlearned MoE LLMs predominantly closer to the bottom right corner, indicating a significant and unacceptable utility drop compared to conventional dense LLMs.

To look into this phenomenon, we begin by performing a careful sanity check on unlearning methods in MoE LLMs and conduct an in-depth analysis of failure cases. Ideally, in MoE LLMs, given an input, the routers should evaluate and direct it to the most relevant experts, with unlearning targeting and erasing the corresponding knowledge in these experts. However, by monitoring expert selection during unlearning, we find that the process often prompts routers to constantly switch the activated experts. This behavior persists even when routers are fixed. As a result, unlearning algorithms create ``short-cuts'', where instead of targeting the most relevant experts, the routers shift to less relevant ones to trick for unlearning loss reduction (\textit{i.e.}, irrelevant experts are unlearned). This leads to substantial drops in model utility (illustrated in \textbf{Fig.\,\ref{fig: teaser}}(b)). %See \textbf{Fig.\,\ref{fig: teaser}}(b) for illustration.

To solve the problem, we propose a novel unlearning framework specifically tailored for MoE LLMs, named {\ours} %, which stands for 
(\underline{S}elected \underline{E}xperts \underline{U}nlearning \underline{F}ramework). {\ours} employs expert attribution to pinpoint the experts most actively involved with the forget set, which is designated as the primary target for unlearning. Unlearning efforts are exclusively focused on this identified expert. Concurrently, an anchor loss is applied to the router to stabilize the active status of the targeted expert throughout the unlearning process. This approach prevents the frequent switching of expert selection, ensuring that unlearning is both focused and controlled. Our contributions are summarized below.

% HM: Based on our observations, we propose an unlearning framework {\ours}, specifically designed for MoE models. Leveraging task specialization in expert system, we constrain the affect of unlearning algorithm on model utility by unlearning task-related experts. Then we further recognize the key feature of MoE architectures, where router layers and expert layers should work collaboratively, to improve the effectiveness of unlearning by manipulating the router layer. On our empirical study, we point out unlearning one expert and its corresponding router layer is enough for unlearning tasks. Our contribution can be summarized as 

$\bullet$ We for the first time identify the unique challenge of unlearning in MoE LLMs. Our analysis elucidates the root causes of observed failures, offering novel insights into how unlearning impacts the routers and experts within an MoE LLM.

$\bullet$ We propose a novel parameter-efficient unlearning framework, {\ours},  for MoE LLMs. {\ours} effectively pinpoints, fixates, and unlearns the most pertinent experts relative to the forget set. {\ours} enjoys high flexibility and works in a plug-in-and-play mode with any existing unlearning methods to boost forget quality, model utility, and efficiency at the same time.

$\bullet$ We conduct extensive experiments to demonstrate the effectiveness of {\ours} across various MoE architectures, MU benchmarks, and unlearning methods. Our results show that when integrated with {\ours}, all tested unlearning methods achieve substantial improvements in model utility up to $35\%$ and concurrently enhance the quality of forgetting with only $0.06\%$ parameters being updated. 
% Furthermore, {\ours} shows strong robustness against jailbreak and relearning-based attacks.

%\vspace{-0.15in}
\section{Related Works}
\vspace{-0.15in}
% \vspace{-0.4cm}
\textbf{Machine Unlearning for LLMs.}
A growing body of research has investigated the problem of unlearning in  LLMs \citep{yao2024machine,lu2022quark,jang2022knowledge,kumar2022privacy,zhang2023forget,pawelczyk2023context,eldan2023whos,ishibashi2023knowledge,yao2023large,maini2024tofu,zhang2024negative,li2024wmdp,wang2024large,jia2024soul,liu2024rethinking,liu2024large,thaker2024guardrail}. These studies have practical applications, such as removing sensitive information \citep{jang2022knowledge,eldan2023whos,wu2023depn}, preventing the generation of harmful or biased content \citep{jang2022knowledge,eldan2023whos,wu2023depn,lu2022quark,yu2023unlearning,yao2023large,liu2024towards}, memorized sequences \citep{jang2022knowledge,barbulescu2024each}, and copyrighted material \citep{eldan2023whos, jang2022knowledge}. To facilitate unlearning, recent methods aim to bypass the need for retraining models from scratch by excluding the forget set containing the targeted data to be removed \citep{ilharco2022editing,liu2022continual,yao2023large,eldan2023whos,jia2024soul,zhang2024negative,li2024wmdp, thaker2024guardrail,liu2024large}. Techniques like task arithmetic also enable efficient model editing through parameter merging \citep{hu2024separate,ilharco2022editing}. Although these methods do not provide exact unlearning akin to full retraining, they remain efficient and effective under empirical unlearning evaluation metrics. Approaches often include model fine-tuning and optimization \citep{liu2022continual,yao2023large,eldan2023whos,jia2024soul,zhang2024negative,li2024wmdp}, or input prompting and in-context learning \citep{thaker2024guardrail,pawelczyk2023context,liu2024large}. Other approaches, such as localization-informed unlearning, identify and locally edit model units (e.g., layers or neurons) closely related to the data or tasks being unlearned \citep{meng2022locating,wu2023depn,wei2024assessing}. %Despite these efforts, studies have shown that forgotten knowledge can often still be extracted from models post-unlearning \citep{patil2023can, liu2024threats, lynch2024eight, shostack2024boy}. However, 
Most existing research has focused on dense LLMs, leaving unlearning in MoE LLMs unexplored. For example, the unlearning of Mixtral-$8\times7B$  discussed in \citet{li2024wmdp}  only examined a single method with ad-hoc adjustments. This work aims to fill this gap by conducting a comprehensive study of various unlearning methods, benchmarks, and MoE models, addressing the specific challenges posed by the MoE architecture.

\textbf{MoE-based LLMs.}
Sparse  MoE models are designed to activate only a subset of expert networks for each input during inference, enabling substantial model scaling with minimal computational overhead \citep{shazeer2017outrageously}. Current %approaches to 
MoE model development can be categorized into two types: training from scratch \citep{fedus2022switch,zoph2022designing,shen2023moduleformer} and building from dense checkpoints \citep{zhang2021moefication,komatsuzaki2022sparse,zhu2024llama}. Over recent years, MoE models have seen key advancements, including improvements in scalability \citep{riquelme2021scaling,kim2021scalable,zhou2022mixture,zoph2022designing}, efficiency optimization \citep{fedus2022switch,lepikhin2020gshard,chowdhery2023palm}, and expert balancing techniques \citep{cong2024prediction,zoph2022st,dai2022stablemoe}.
The implementation of transformer-based MoE models has been integrated into LLMs, significantly enhancing inference efficiency \citep{jiang2024mixtral,dai2024deepseekmoe,grok1,hong2024reference,abdin2024phi,lieber2024jamba,yang2024qwen2,zhu2024llama,databricks_dbrx_2024}. For example, DeepSeekMoE \citep{dai2024deepseekmoe} improves expert specialization by segmenting experts into smaller subsets for flexible activation, while isolating shared experts to reduce redundancy and capture common knowledge. Similarly, Qwen1.5-MoE \citep{qwen_moe} partitions a standard FFN layer into smaller segments to create multiple experts, introducing a fine-grained routing mechanism that enables Qwen1.5-MoE to match the performance of 7B models with only one-third of parameters activated.
Despite the efficiency gains provided by MoE's dynamic routing system, existing research highlights additional challenges compared to traditional dense models, including unstable training \citep{zoph2022designing,dai2022stablemoe}, robustness issues \citep{zhang2023robust,puigcerver2022adversarial}, and complications in parallel deployment \citep{hwang2023tutel,gale2023megablocks}. In this work, we show that the root cause of the ineffectiveness of existing unlearning methods for MoE LLMs also stems from the dynamic routing system.

\section{Preliminaries}
\vspace*{-0.05in}
\label{sec: exploration}
In this section, %we start by presenting the mathematical formulation of LLM unlearning. The lack of exploration on MoE LLM unlearning inspires us to investigate whether existing unlearning methods keep effective in these models. 
we present our pilot study to reveal that unlearning methods designed for conventional LLMs are \textit{ineffective} in unlearning MoE LLMs. 

\textbf{Preliminaries on MoE LLM unlearning.} Based on the generic formulation %of LLM unlearning 
outlined in \citet{liu2024rethinking}, the task of LLM unlearning is to eliminate the influence of a specific ‘unlearning target’–whether it is related to data, knowledge, or model capabilities–from a pretrained LLM (denoted by $\btheta_o$). The unlearning target is typically defined by a forget set $\mathcal{D}_f$, which contains the information or knowledge to be removed. To ensure the model retains its generation ability (\textit{i.e.}, utility) after unlearning, a retain set $\mathcal{D}_r$ is introduced, consisting of data unrelated to the unlearning target. With this setup, the LLM unlearning problem is usually formed as a regularized optimization problem, finetuned from $\btheta_o$ using both the forget set $\mathcal{D}_f$ and the retain set $\mathcal{D}_r$:
\vspace{-0.05in}
\begin{align}
    \min_{\btheta} \ell_f(\btheta; \mathcal{D}_f) + \lambda \ell_r(\btheta; \mathcal{D}_r).
\end{align} \vspace{-1.5em}

Here, $\btheta$ represents the model parameters to be updated during unlearning, $\ell_f$ and $\ell_r$ denote the forget loss and retain loss, respectively, with $\lambda \geq 0$ serving as a regularization parameter to balance between unlearning and preserving utility. 

%\SL{[The following section mixed too many maths and texts and does not look well.]}
Next, we provide a brief introduction to how the routing system operates in the MoE LLM architecture.
In MoE LLMs, \textit{e.g.}, DeepSeek-v2-Lite \citep{liu2024deepseek}, the feed-forward networks (FFNs) of Transformers are split into multiple experts and activated by the output of the router in front of the expert layers, see Fig.\,\ref{fig: teaser}(b) for illustration. 
In the $l$-th layer, given the input $\mathbf{u}_t^{(l)}$ corresponding to the $t$-th token, router layers calculate the score of each token and assign them to the top-$K$ experts:
\vspace*{-0.5em}
\begin{align*} 
s_{i,t}^{(l)} &= \text{Softmax}(\text{Router}(\mathbf{u}_t^{(l)})) \\
g_{i,t}^{(l)} &=
\begin{cases}
s_{i,t}^{(l)} & \text{if } s_{i,t}^{(l)} \in \text{Top}K(\{s_{k,t}^{(l)} \mid 1 \leq k \leq N\}) \\
0 & \text{otherwise}
\end{cases}
\end{align*}
\vspace*{-1em}

% \begin{align*}
% g_{i,t}^{(l)} &=
% \begin{cases}
% s_{i,t}^{(l)} & \text{if } s_{i,t}^{(l)} \in \text{Top}K(\{s_{k,t}^{(l)} \mid 1 \leq k \leq N\}) \\
% 0 & \text{otherwise}
% \end{cases}
% \end{align*}

%
Here, $\text{Router}(\cdot)$ denotes the router layer, $s_{i,t}$ is the token-to-expert affinity, $\text{Top}K(\cdot)$ selects the highest $K$ value in the set,  $N$ is the number of experts, and $g_{i,t}^{(l)}$ is the score assigned by router for the $i$-th expert.
Then, the hidden state $\mathbf{h'}_t^{(l)}$ of FFNs can be calculated as: $\mathbf{h'}_t^{(l)} = \mathbf{u}_t^{(l)} + \sum_{i=1}^N g_{i,t}^{(l)} \, \text{FFN}^{(l)}_i(\mathbf{u}_t)$, where  $\text{FFN}^{(l)}_i(\cdot)$ denotes the $i$-th expert.
Then, $\mathbf{h'}_t^{(l)}$ is sent to the next layer of Transformer blocks for further processing.

\begin{table}[t]
%    \vspace*{-0.4em}
    \centering
    \caption{ {Unlearning performance of GA when controlling tunable parameters in MoE LLMs.}}
    \vspace*{-0.5em}
    \resizebox{0.9\linewidth}{!}{
    \begin{tabular}{cc|ccc}
    \toprule[1pt]
    \midrule
    \multicolumn{2}{c}{Tunable Module} & Forget Efficacy $\downarrow$ & Utility $\uparrow$ \\
    
    \midrule    
     \multirow{4}{*}{Qwen}&Original & 0.4192 & 0.5979\\
     % \midrule
      &Experts \& Router & 0.2953   & 0.3393  \\
      &Routers Only & 0.2526  & 0.2977  \\
      &Experts Only & 0.2536 & 0.3242  \\     
%     Experts \& Router & 0.2953 (0.1239$\uparrow$) & 0.3393 (0.2586$\downarrow$)\\
%     Routers Only & 0.2526 (0.1666$\uparrow$)& 0.2977 (0.3002$\downarrow$)\\
%     Experts Only & 0.2536 (0.1656$\uparrow$)& 0.3242 (0.2737$\downarrow$)\\
     \midrule
       
     \multirow{4}{*}{DeepSeek}&Original & 0.3804 & 0.5500\\
     % \midrule
    &Routers \& Expert & 0.2457  & 0.3145  \\
     &Routers Only & 0.2375  & 0.3315  \\
     &Experts Only & 0.2601  & 0.3435  \\
%     Routers \& Expert & 0.2457 (0.1347$\uparrow$)& 0.3145 (0.2355$\downarrow$)\\
%     Routers Only & 0.2375 (0.1429$\uparrow$)& 0.3315 (0.2188$\downarrow$)\\
%     Experts Only & 0.2601 (0.1203$\uparrow$)& 0.3435 (0.2065$\downarrow$)\\
     \midrule
     \bottomrule[1pt]
    \end{tabular}}
    \vspace*{-1em}
    \label{tab: three_settings}
    %\vspace*{-0.2em}
\end{table}

\textbf{Unlearning for MoE LLM is not trivial: a pilot study.} The goal of unlearning is twofold: (1) to ensure the model forgets the targeted information and knowledge stored in $\mathcal{D}_f$, and (2) to preserve the model utility without significant degradation. Our pilot study reveals that the special routing system in MoE LLMs introduces additional challenges to unlearning, rendering existing methods ineffective.
We applied
%\textbf{Fig.\,\ref{fig: teaser}(a)} empirically justifies that the direct application of existing unlearning method is problematic. We evaluated 
four widely used LLM unlearning methods: {\ga} (Gradient Ascent) \citep{eldan2023whos}, {\gdiff} (Gradient Difference) \citep{maini2024tofu}, {\npo} (Negative Preference Optimization) \citep{zhang2024negative}, and {\rmu} (Representation Misdirection for Unlearning) \citep{li2024wmdp} with the {\wmdp} benchmark \citep{li2024wmdp} on two MoE LLMs, Qwen1.5-MoE \citep{qwen_moe} and DeepSeek-V2-Lite \citep{liu2024deepseek}, as well as two dense LLMs for reference, LLaMA3-8B \citep{dubey2024llama} and Phi-3.5-mini-instruct \citep{abdin2024phi}, where the task aims to unlearn hazardous knowledge in LLMs. In \textbf{Fig.\,\ref{fig: teaser}(a)}, to ease the comparison, we report the forget quality (performance drop on the forget test set, where higher is better) against retain quality (performance drop on the MMLU \citep{hendrycks2020measuring} utility benchmark, where lower is better).
Each data point represents the best result of a model-method combination with hyper-parameter tuning, with ideal performance located near the top left corner, signifying high unlearning effectiveness with minimal impact on model utility. As we can see, most MoE LLM data points cluster in the lower right, indicating severe utility drops and poor unlearning performance compared to dense models. In Fig.\,\ref{fig: teaser}(a), all model parameters (including routers and experts) are involved in unlearning. To ensure that these poor results are not due to improper parameter settings, \textbf{Tab.\,\ref{tab: three_settings}} presents additional experiments using two other parameter configurations (routers-only and experts-only) for {\ga}, yet no significant improvements are observed in either forget or retain quality (more than 20\% utility drop). The results above imply the problem of MoE LLM unlearning is more challenging and far from trivial, even if LLM unlearning is well-studied.
\section{Our Proposal: \ours}
%{Selected Experts Unlearning Framework} (\ours)}
\label{sec: method}
\vspace{-0.1in}
% \SL{Why do you call UoE? Better to recall or provide the full name above.}
In this section, we delve into the failure cases highlighted in Sec.\,\ref{sec: exploration} by analyzing the behavior of routers and their expert selection patterns. We then identify two primary root causes underlying the poor unlearning performance in MoE LLMs. Based on these insights, we introduce {\ours}, a new unlearning paradigm designed to achieve controllable and effective unlearning for MoE LLMs.

\begin{figure}
    \centering
    \begin{tabular}{ccc}
         \hspace*{-4mm} \includegraphics[width=0.35\linewidth]{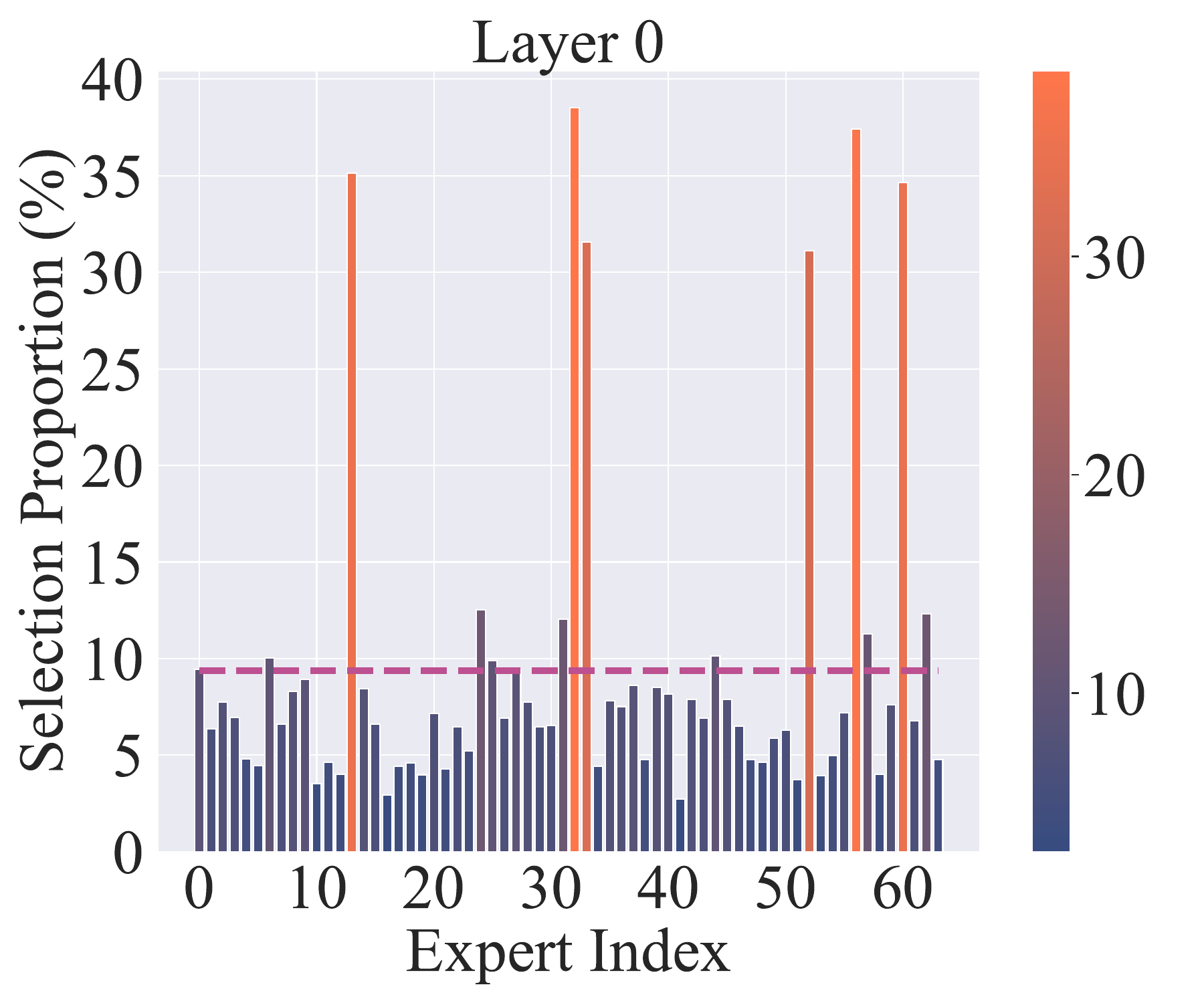} 
         % & \includegraphics[width=0.25\linewidth]{figs/layer_distribution/expert_dist_layer_10.pdf} 
         & \hspace*{-7mm} \includegraphics[width=0.35\linewidth]{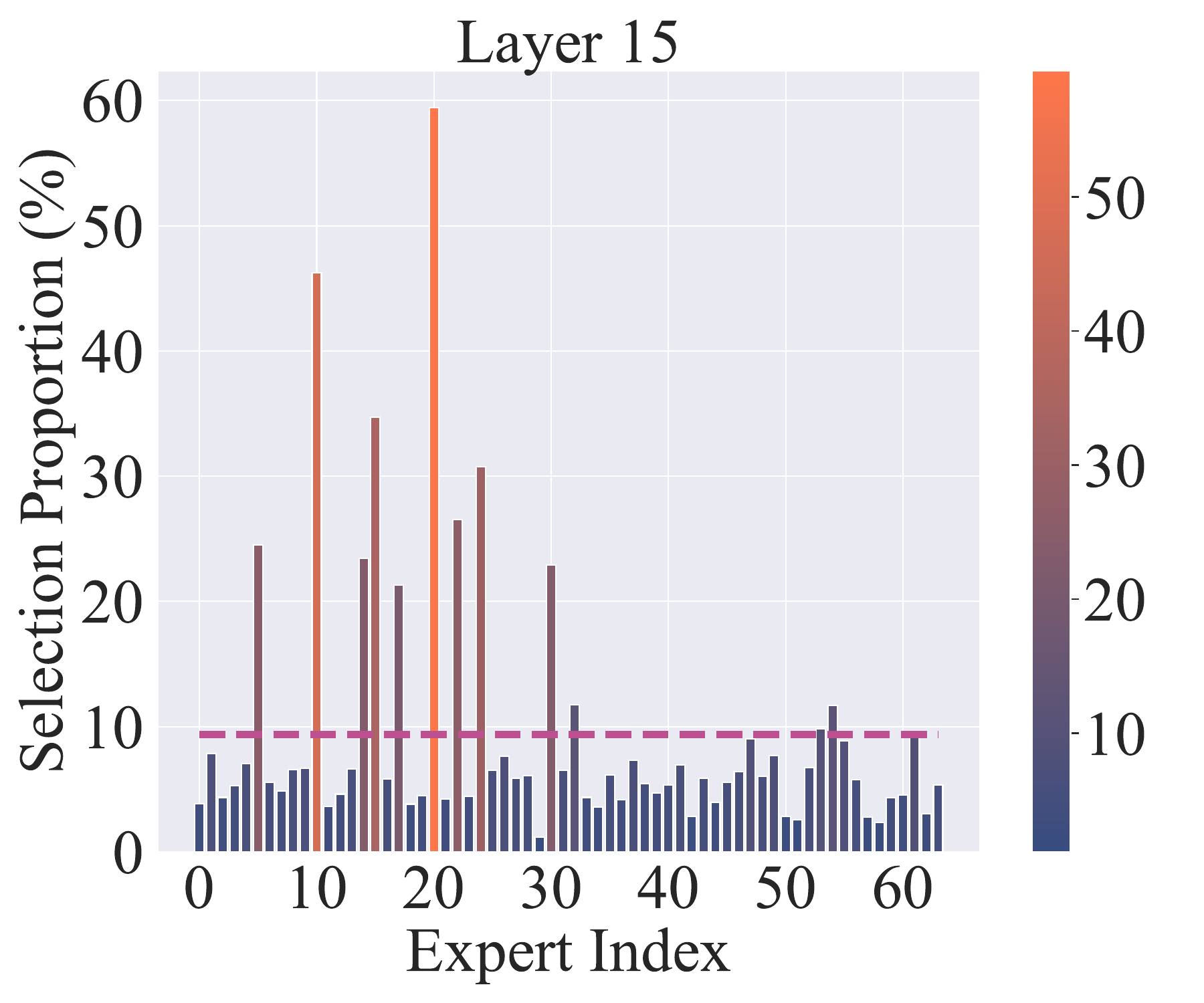} 
         & \hspace*{-7mm} \includegraphics[width=0.35\linewidth]{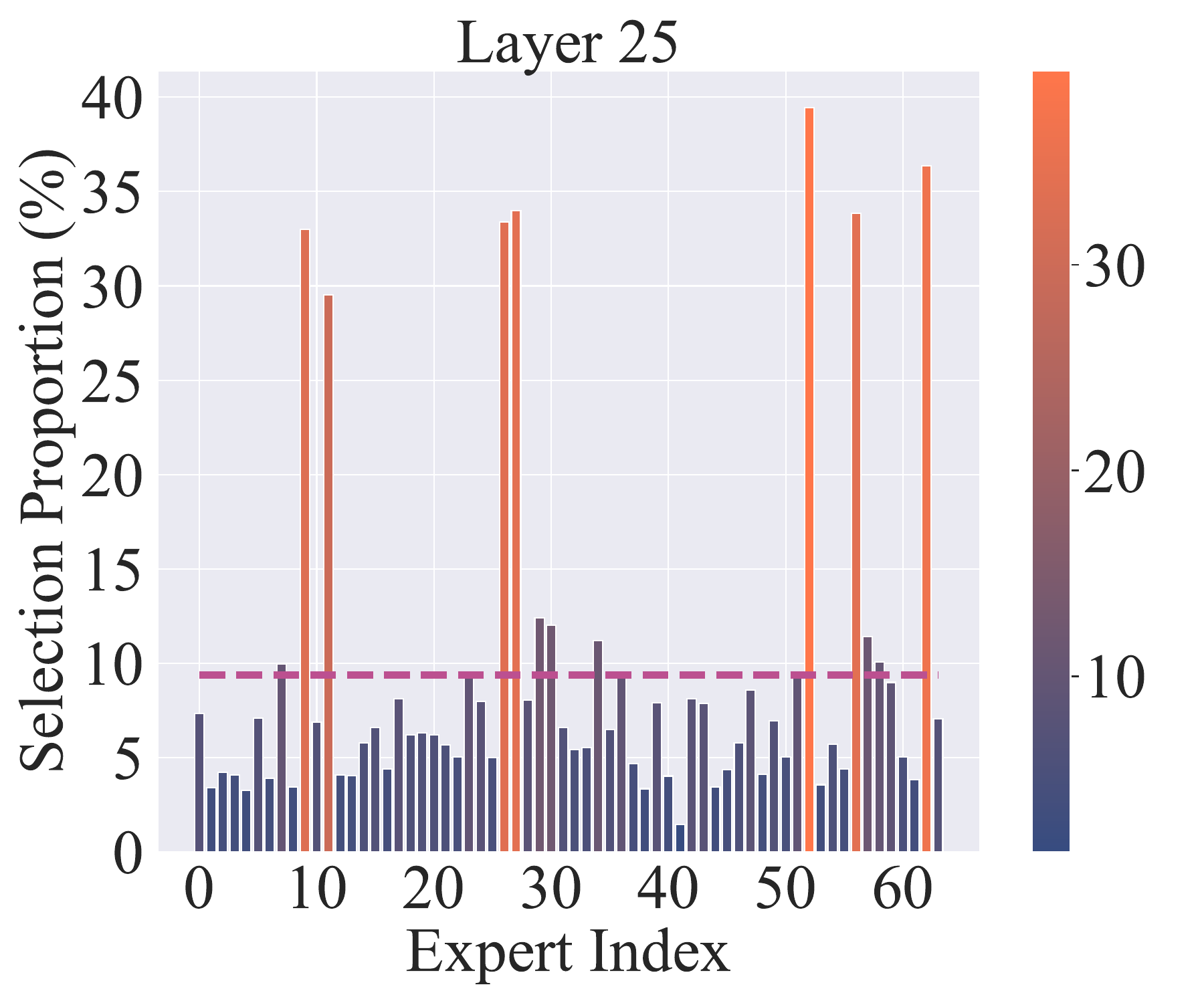}
    \end{tabular}
    \vspace*{-1em}
    \caption{\footnotesize{Proportion of tokens assigned to each expert of the pre-trained DeepSeek-v2-Lite ($K$=6 in Top$k$) with  samples from  {\wmdp} forget benchmark \citep{li2024wmdp}, in different model layers. The dashed horizontal line marks 6/64, \textit{i.e.}, the proportion expected with uniform expert selection. The expert selection distribution clearly follows a long-tailed pattern when the input is sampled from a topic within a narrow scope.}}
    \label{fig: expert_selection_dist}
    %\vspace*{-0.5em}
\end{figure}

\textbf{Uncovering the root cause: `short-cut' in MoE LLM unlearning and expert selection shift.} In order to fully understand the failure cases of MoE LLM unlearning, we begin by inspecting and monitoring the expert selection pattern of the unlearned model. In \textbf{Fig.\,\ref{fig: expert_selection_dist}}, we show the proportion of tokens assigned to each selected expert on the data samples from {\wmdp}  forget dataset \citep{li2024wmdp}. For the input of a specific topic, a small portion of experts (around 6  to 9  out of 64 experts) % in Fig.\,\ref{fig: expert_selection_dist} when selecting 6 experts each time) 
were assigned with the majority of the tokens in each layer, which was also confirmed in \citet{wang2024let}. Thus, we have the following insight:

\begin{center}
\vspace*{-1.8em}
	\setlength\fboxrule{0.5pt}
	\noindent\fcolorbox{black}[rgb]{0.99,0.99,0.99}{\begin{minipage}{0.99\columnwidth} \small
        {\bf Insight}\,1: \textit{For the inference related to a certain topic within a narrow scope (\textit{e.g.,} the forget set of an unlearning task), expert selection by MoE routers follows a long-tailed distribution, with only a few experts being activated significantly more frequently than others.}
	\end{minipage}}
\end{center}

\begin{figure}[t]
    \vspace*{-0.2em}
    \centering
    \begin{tabular}{cc}
         \hspace*{-4mm} \includegraphics[width=.34\linewidth]{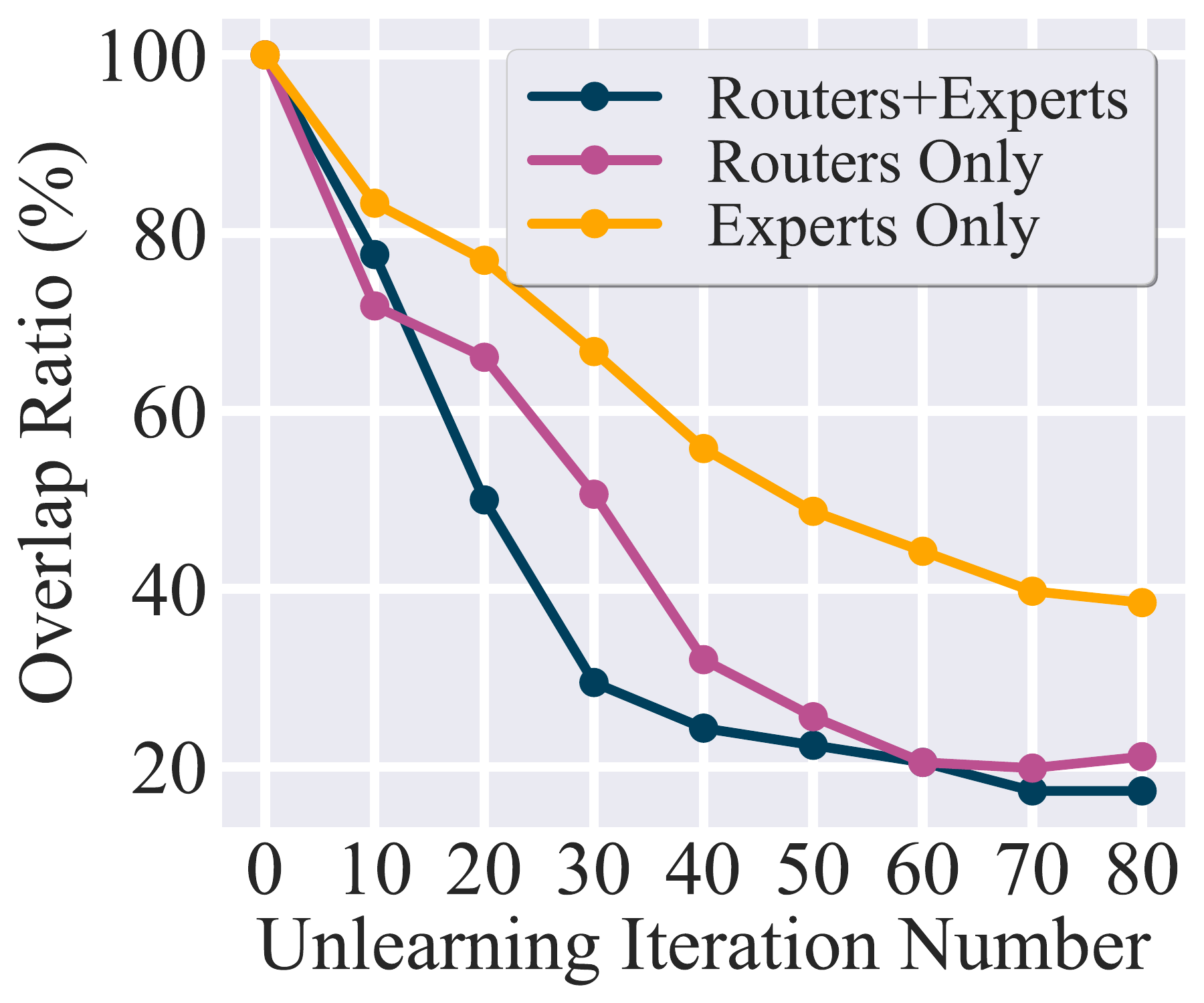} 
         & \hspace*{-6mm} \quad \includegraphics[width=.34\linewidth]{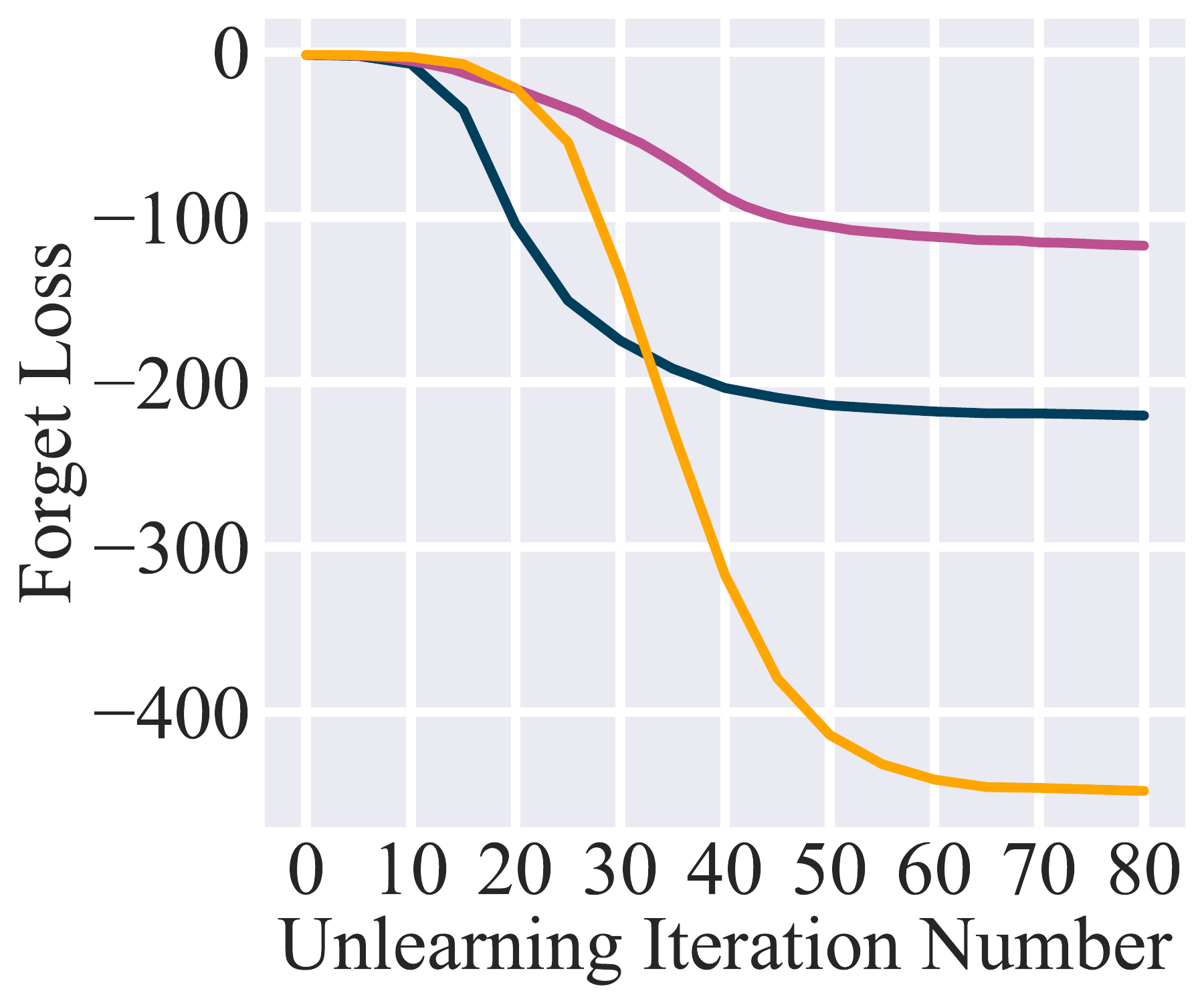} \\
      %   \hspace*{-4mm} (a)     & \hspace*{-6mm} (b)
    \end{tabular}
    \vspace*{-1em}
    \caption{\footnotesize (left) Overlap ratio of selected experts between the original pretrained model and the unlearned model with different unlearning iterations using {\ga} on {\wmdp} benchmark. (right) Forget loss vs. the number of unlearning iterations, when controlling  parameters to unlearn in MoE LLM.}
    \label{fig: overlap_ratio}
    \vspace*{-1em}
\end{figure}

Based on the insight above, we define the frequently activated experts as \textbf{topic-\textit{target} experts}, and the others as \textbf{\textit{non-target}}. Thus, by eliminating the knowledge stored in these target experts, MoE LLM unlearning can be achieved more effectively. 

Next, we examine how the expert selection pattern evolves during unlearning. Specifically, we track the average expert selection overlap ratio across all layers between the unlearned model at different stages and the original pretrained model, when processing the forget set. The results, shown in \textbf{Fig.\,\ref{fig: overlap_ratio} (a)}, reveal a steady decline in the overlap ratio as unlearning progresses, indicating that previously selected target experts are gradually replaced by non-target ones that do not contain the target knowledge. This shift persists even when routers are fixed, as unlearning can still indirectly influence router selection: a router's decision at one layer depends on the output of the previous layer, which may have been affected by an updated expert of this previous layer in unlearning. Meantime, we observe a consistent reduction in forget loss, as shown in \textbf{Fig.\,\ref{fig: overlap_ratio} (b)}. Thus, we can derive the following insight:
\vspace{-0.2in}
\begin{center}%\vspace{-0.3in}
	\setlength\fboxrule{0.5pt}
	\noindent\fcolorbox{black}[rgb]{0.99,0.99,0.99}{\begin{minipage}{0.99\columnwidth} \small
        {\bf Insight\,2}: \textit{Existing unlearning methods tend to prompt routers to shift  selection from target to non-target experts unintentionally. This creates unlearning `shortcuts' in expert selection to trick for low forget loss and lead to fake unlearning}.
	\end{minipage}}
\end{center}
\vspace{-0.02in}

As unlearning proceeds, non-target experts are more frequently activated to handle samples related to the unlearning target, thereby being forced to participate in the unlearning task, even though they did not contain the intended target knowledge. Meanwhile, the true objective of unlearning, \textit{i.e.,} the target experts, remain hidden out of the reach of the forward propagation.
Existing literature \citep{liu2024rethinking} has already demonstrated that forcing unlearning models that do not contain knowledge related to the unlearning target can cause a significant drop in model utility. 
This accounts for the sharp decline in model utility observed in Sec.\,\ref{sec: exploration}, which leads to the following insight:
\vspace{-0.05in}

\begin{center}
    \vspace*{-1em}
	\setlength\fboxrule{0.5pt}
	\noindent\fcolorbox{black}[rgb]{0.99,0.99,0.99}{\begin{minipage}{0.99\columnwidth} \small
        {\bf Insight\,3}: \textit{The sharp degradation in model utility during MoE LLM unlearning is primarily due to excessive unlearning applied to non-target experts caused by expert selection shift.}
	\end{minipage}}
\end{center}

% \begin{wrapfigure}{r}{0.5\linewidth}
%     \vspace*{-1em}
%     \centering
%     \includegraphics[width=\linewidth]{figs/place_holder.pdf}
%     \vspace*{-1.2em}
%     \caption{\footnotesize An illustration of the different components in {\ours}.}
%     \label{fig: method_illustration}
%     \vspace*{-1em}
% \end{wrapfigure}
\textbf{{\ours} for effective MoE LLM unlearning.} 
% \SL{Please use consistent paragraph title format.}
As discussed earlier, a new paradigm tailored for MoE LLM unlearning is urgently needed to address the challenges of unintentional expert selection shifts in routers and excessive unlearning of non-target experts. Therefore, we propose a framework that (1) identifies the most relevant target experts, (2) ensures that these target experts remain highly activated throughout the unlearning process to avoid selection shifts, and (3) limits the impact of unlearning on non-target experts. Spurred by these, we introduce {\ours}, where unlearning is confined to $M$ most relevant target experts. We refer the readers to {Alg.\,\ref{algorithm}} for an illustration of {\ours}. 

This approach starts with an  expert attribution process to accurately identify the most $M$ 
 relevant experts for the unlearning task (step 1-3). Then, the gradient computation selected experts $e_{M}$ and their corresponding routers $\text{R}_{e_{M}}$ are enabled (step 4), while other parameters are frozen. Step 5 performs unlearning using any unlearning approach, as our framework is flexible. For example, gradient ascent can be applied with our defined loss functions.  Next, we present the details of the expert attribution process and define the anchor loss function. 

%\begin{wrapfigure}{r}{0.5\textwidth}
  \begin{minipage}{0.45\textwidth}
  \vspace*{-0.5em}
    \begin{algorithm}[H]
    \caption{\small{\ours} Unlearning Algorithm}
    \label{algorithm}
    \begin{algorithmic}[1]
    \footnotesize 
    \ENSURE{Unlearned model $\btheta_u$}
    \REQUIRE{Pretrained model $\btheta_o$, forget set $\mathcal{D}_f$, retain set $\mathcal{D}_r$, 
  
  \noindent\hspace{-1.9\algorithmicindent} \textbf{Setup:} Retain loss $\ell_r$, forget loss $\ell_f$, anchor loss $L_{\text{anchor}}$}, {the number of experts to select} $M$ %Selected Number $M$

        \STATE $\mathcal{D}_{s}\gets$ Sample\_Subset($\mathcal{D}_f$)
        \STATE $s\gets$ Record\_Affinity\_Score($\btheta_o, D_{s}$)
        \STATE $e_{M}\gets$Ranking\_And\_Select($s, M$)
        \STATE Activate\_Expert\_And\_Router($\btheta_o, e_{M}, $$\text{R}_{e_{M}}$)
        %\vspace{-0.1in}
        \STATE $\btheta_u\gets$Unlearn$(\btheta_o,\ell_f(\mathcal{D}_f),\ell_r(\mathcal{D}_r),L_{\text{anchor}})$
        %\vspace{-0.15in}
        \STATE Return $\btheta_u$
    \end{algorithmic}
    \end{algorithm}
  \end{minipage}
  %\vspace*{-0.5em}
%\end{wrapfigure}

\vspace{+0.1in}
\ding{70} \textbf{Expert attribution.} While the token assignment ratio for each expert (shown in Fig.\,\ref{fig: expert_selection_dist}), can serve as a basic attribution metric, it overlooks finer details that are important for precise comparisons{, due to the hidden states in each layer summed by weighted average}. To address this, we adopt a gating score-based task affinity calculation method from \citep{wang2024let}. Specifically, the affinity score for the $i$-th expert $e_i^{(l)}$ in the $l$-th layer of an MoE LLM is defined as:
\vspace*{-0.05in}
\begin{equation}\vspace*{-0.05in}
    s_i^{(l)}=\frac{1}{Z}\sum_{j=1}^{Z} \frac{1}{L_j} \sum_{t=1}^{L_j} g_{i, t}^{(l)}
    \label{eq: affinity_score}\vspace*{-0.05in}
\end{equation}
where $Z$ is size of the calibration dataset used for expert attribution, $L_j$ represents the length of the $j$-th input sequence $\bx_j$, and $g_{i,t}^{(l)}$ is the probability score assigned to expert $\mathbf{e}_i^{(l)}$ for the $t$-th token. Following \citet{wang2024let}, the attribution data can be a subset universally sampled from the original forget set. We find that a subset containing over 100,000 tokens is robust enough to select the most relevant experts for an unlearning task. For each layer, we rank the experts based on their affinity score and then finally select the top $M$ experts as the target expert for unlearning ($e_M$  in Algo. 1).

\ding{70} \textbf{Router anchor loss.} A key challenge in unlearning is the expert selection shift, where the true target experts are hidden by the routers, while less relevant experts are activated during inference and inadvertently involved in the unlearning process. To mitigate this, we propose the router anchor loss, which encourages the previously identified target expert to remain consistently activated throughout unlearning. The loss is formulated as:
\vspace*{-0.3em}
\begin{equation}\vspace{-0.05in}
    L_{\text{anchor}}^{(l)} = \| \mathbf{g}^{(l)} - [a_1^{(l)}, a_2^{(l)}, \dots, a^{(l)}_{E^{(l)}}] \|_2^2, \,\, 
\end{equation}
where $E^{(l)}$ is the total number of experts in the $l$-th layer, $ \mathbf{g}^{(l)}=[g^{(l)}_1,g^{(l)}_2,\dots,g^{(l)}_i]$ is the output of router, and $a_i^{(l)} = 1$ if the $i$-th expert is identified as the target expert, otherwise $a_i^{(l)} = 0$. {The unlearning loss can then be formularized as:}
\vspace*{-0.05in}
\begin{equation}\label{eq:loss} \vspace*{-0.05in}
    \min_{\btheta} \ell_f(\btheta; \mathcal{D}_f) + \lambda \ell_r(\btheta; \mathcal{D}_r)+\alpha L_{\text{anchor}}^{(l)},
    \vspace*{-0.05in}
\end{equation}
where $\alpha$  controls the strength of anchor loss. Its sensitivity is analyzed in Appendix
%. The sensitivity analysis of $\alpha$ is provided in
Sec.~\ref{sec:hyperparameter}.

\ding{70} \textbf{Selection of top $M$ experts.}
When forming $e_M$ of the top $M$ experts, there are two approaches: 1) selecting the top $M$  experts from all experts across all layers based on the affinity score $s_i^{(l)}$ in Eq.\ref{eq: affinity_score}; and 2) to mitigate selection shift from previous layers, another approach is to choose the top $M$ experts from the same layer. We examined both approaches under different settings $M$=1,3,6, and present the results in Tab. \ref{tab:NumExpert}. We observe that unlearning a single expert ($M$=1) yields better performance than unlearning multiple experts, regardless of whether they come from the same layer or different layers. 
This trend of single-expert unlearning yielding the best performance is also observed across other unlearning tasks (see Tab.~\ref{tab:NumExpert_qwen_rwku} in Appendix).
%This trend is not limited to using GA for unlearning on the WMDP benchmark, we examined   additional unlearning tasks and found that single-expert unlearning ($M$=1) consistently achieves the best performance across different scenarios.} 
This suggests: 
\begin{center}
    \vspace*{-1.5em}
	\setlength\fboxrule{0.5pt}
	\noindent\fcolorbox{black}[rgb]{0.99,0.99,0.99}{\begin{minipage}{0.99\columnwidth}\small
        {\bf Insight\,4}: \emph{Unlearning top-1  expert is the most effective.}
	\end{minipage}}
\end{center}

% \usepackage{multirow}
\begin{table}[t]
\centering
\caption{\footnotesize Model utility (UT$\uparrow$) comparison at the same level of forget efficacy (FE$\approx0.25$),
when the top $M$ experts from either the same layer or different layers in DeepSeek are unlearned using {\ga} on {\wmdp} benchmark, also when 4 shared experts are included.}
\vspace{-0.1in}
\label{tab:NumExpert}

\scalebox{0.72}{
\begin{tabular}{l|llcc}
\toprule
\multicolumn{1}{l|}{Selected experts} & \multicolumn{1}{c}{Top-1} & \multicolumn{1}{c}{Top-3} & Top-6 & Top-1+4-shared \\
\midrule
\multirow{1}{*}{\centering Same layer}
    & \multirow{2}{*}{0.5100} & 0.4856 & 0.4652 & 0.3554 \\
   \multicolumn{1}{l|}{Different layers}    &                         & 0.2852 & 0.2567 & --     \\
\bottomrule
\end{tabular}
}
\vspace{-0.15in}
\end{table}

%While we have successfully identified the most relevant experts, the optimal number of experts for unlearning and the best strategy for selecting them remain unclear. There are two approaches to choosing top experts: selecting experts from the same layer or selecting the top-1 experts from different layers. As shown in Tab.~\ref{tab:NumExpert}, unlearning a single expert achieves better performance compared to unlearning multiple experts, whether they are from the same layer or different layers. Moreover, unlearning experts from different layers results in a significant performance drop.

From Tab.~\ref{tab:NumExpert}, we also observe that unlearning multiple experts across different layers leads to a substantial performance decline. To further analyze the Insight 4, %and observations through gradient updates in unlearning, 
let the total gradient update during unlearning   be:
\(
\Delta W = \sum_{i \in e_{M}} \lambda_i \nabla \mathcal{L}_i,   
\)
where \( e_{M} \) is the set of selected experts being unlearned, \( \lambda_i \) denotes their contribution weight, and \( \nabla \mathcal{L}_i \) is their corresponding gradient update  in Eq. (\ref{eq:loss}). When only the top-1 expert is selected for unlearning, the modification to the weights remains minimal, ensuring low gradient interference. For multiple experts within the same layer, the gradient updates may partially cancel out, leading to moderate disruption. However, for multiple experts across different layers, the gradient updates affect distinct feature hierarchies, resulting in an unstable gradient flow and widespread model disruption.

This analysis also explains the deficiency of unlearning shared experts. In a given layer, shared experts are activated for all tokens, making them intuitively suitable targets for unlearning. However, Tab.~\ref{tab:NumExpert} shows that unlearning the top-1 expert along with 4 shared experts causes a greater utility drop than unlearning top-6  experts in the same layer. Shared experts influence a broader range of token representations, so making them active for unlearning  triggers high-magnitude gradient updates across multiple pathways. Also, since shared experts consolidate common knowledge across diverse contexts~\cite{liu2024deepseek}, their modification disrupts the model more severely, making them suboptimal for unlearning.

\begin{table*}[t!]
\centering
\caption{\footnotesize Performance comparison of existing unlearning methods equipped w/ and w/o {\ours} on {\wmdp} \citep{li2024wmdp} and {\rwku} \cite{jin2024rwku} benchmarks on two MoE LLMs, namely Qwen1.5-MoE-A2.7B-Chat (Qwen) \cite{qwen_moe} and DeepSeek-V2-Lite (DeepSeek) \citep{dai2024deepseekmoe}. %The $\uparrow$ and $\downarrow$ symbols denote metrics where higher/lower values are better. 
Additionally, a group of baselines applying PEFT (LoRA and ESFT) on GA is included to evaluate our method's effectiveness in selecting a suitable subset of parameters for unlearning, along with a baseline using random expert selection with RMU.
The occurrence of significant utility increase (over $5\%$ increase in UT compared to without SEUF) are marked in \congrat{green}. }
\vspace*{-.5em}
\resizebox{0.7\linewidth}{!}{
\begin{tabular}{c|cc|cc|cc|cc}
\toprule[1pt]
\midrule
\multirow{2}{*}{\textbf{Method}}  & \multicolumn{2}{c|}{\textbf{Qwen ({\wmdp})}} & \multicolumn{2}{c|}{\textbf{DeepSeek ({\wmdp})}}& \multicolumn{2}{c|}{\textbf{Qwen ({\rwku})}} & \multicolumn{2}{c}{\textbf{DeepSeek ({\rwku})}} \\ 
               &  \textbf{FE$\downarrow$} & \textbf{UT$\uparrow$} & \textbf{FE$\downarrow$} & \textbf{UT$\uparrow$} & \textbf{FE$\downarrow$} & \textbf{UT$\uparrow$}& \textbf{FE$\downarrow$} & \textbf{UT$\uparrow$} \\ \midrule
\textbf{Pretrained}          & $0.4192$        & $0.5979$         & $0.3804$        & $0.5548$     & $0.4243$        & $0.5979$         & $0.5376$        & $0.5548$     \\ \midrule
\textbf{\ga}   & $0.2953$        & {$0.3393$}         & $0.2457$        & {$0.3145$}    & $0.0078$        & {$0.4849$}         & $0.0839$        & $0.5195$      \\
\textbf{\ga+{\ours}}     &      $0.2987$   &     \congrat{$0.5012$}     & $0.2700$        & \congrat{$0.5100$}     &      $0.0060$   &     \congrat{$0.5709$}    & $0.0000$        & $0.5485$       \\
\midrule
\textbf{\gdiff}      & $0.2964$        & {$0.2965$}         & $0.2898$        & {$0.3929$}      & $0.0700$        & $0.5296$         & $0.1901$        & {$0.3495$} \\ 
\textbf{\gdiff+{\ours}}    &  $0.2445$        & \congrat{$0.5295$}         & $0.2677$        & \congrat{$0.4895$}    & $0.0010$        & \congrat{$0.5987$}        &    $0.0000$     &    \congrat{$0.5253$}     \\\midrule
\textbf{\npo}   & $0.3447$       & {$0.4612$}        & $0.3200$             & $0.4700$       & $0.0000$       & {$0.3718$}        & $0.0970$             & $0.5388$   \\
\textbf{\npo+{\ours}}   & $0.3200$        & \congrat{$0.5468$}         & $0.2898$       & $0.4790$     & $0.0020 $      & \congrat{$0.5428$}        & $0.0000$             & $0.5479$  \\\midrule
\textbf{\rmu}      & $0.2612$        & {$0.3560$}           & $0.2530$ & {$0.4540$} & $0.0200$        & {$0.2420$}        & $0.0010$ & $0.5109$  \\
\textbf{\rmu+{\ours}}    & $0.2536 $      & \congrat{$0.5351$}           & $0.2859$ & \congrat{$0.5424$} & $0.0723$        & \congrat{$0.5975$}           & $0.0130$ & $0.5388$ \\
\midrule
\midrule
% \bottomrule[1pt]
\textbf{GA+LoRA} &0.2459&{0.2689}&0.2657&{0.2295}&$0.0000$&{$0.2689$}&0.0000&{0.2302}\\
\textbf{GA+ESFT} &0.3145&{0.4514}&0.2737&0.5108&0.001&{0.4433}&0.0200&0.5001\\
\midrule
\midrule
% \bottomrule[1pt]
\textbf{\rmu}+\textbf{Random} &0.3505&0.5947&0.2722&0.5183& 0.2110&0.5924& 0.1176&0.5182\\
\midrule
\bottomrule[1.5pt]
\end{tabular}
}
\label{tab: main_results}
%\vspace*{-1.2em}
\end{table*}

 \vspace{-0.1in}
\section{Evaluation Experiments}
\label{sec: experiments}
 \vspace{-0.1in}
To demonstrate the effectiveness of our proposed method, we evaluate and compare it against different baselines on two widely accepted LLM unlearning benchmarks: {\wmdp} \citep{li2024wmdp} and {\rwku} \citep{jin2024rwku}.  The detailed experimental setup, such as unlearning tasks, datasets selection, targeted MoE models, unlearning baselines and hyper-parameter setting, is provided in Appendix Sec.~\ref{Sec: setup},  due to space limitation.  We next present results of several key experiments. 

\ding{70}  \textbf{Effectiveness of {\ours} across benchmarks and unlearning methods.}
%in preserving model utility and unlearning efficacy.} 
In \textbf{Tab.\,\ref{tab: main_results}}, we present the FE (forget efficacy) and UT (utility)  of our proposed {\ours} when integrating different unlearning methods {\ga}, {\gdiff}, {\npo}, and {\rmu}. 
In this evaluation, {\ours} selects only the \textbf{top-1 expert} for unlearning.
There are two notable findings. 
\underline{First}, {\ours} effectively enhances unlearning, either by further reducing FE or maintaining a similar level compared to   baselines without SEUF.
\underline{Second}, {\ours}  consistently improves model utility (UT) across all tested methods. Notably, for methods where UT drops by more than 10\% (compared to the pretrained model), highlighted in red,  {\ours}   mitigates the decline.
For example, the utility of {\ga} on Qwen for the WMDP task drops from $0.5979$ to $0.3393$, but with {\ours}, the utility improves to $0.5012$,
This demonstrates  {\ours}'s effectiveness in \textbf{balancing unlearning performance and model retention}.
%one of the most notable findings is that 
%{\ours} significantly improves model utility (\textbf{UT}) across all tested methods. For instance, when applied to baseline methods like {\ga}, {\gdiff}, and {\rmu}, {\ours} consistently mitigates the severe utility drops (greater than $10\%$) that occur with the unmodified methods. This is particularly evident in scenarios where baseline methods without {\ours} exhibit drastic performance degradation in model utility (highlighted in \warn{red}), while the same methods paired with {\ours} show substantial recovery. For example, the utility of {\ga} on Qwen for the WMDP task drops from $0.5979$ to $0.3393$, but with {\ours}, the utility improves to $0.5012$, restoring much of the lost performance. Similarly, {\gdiff} on {\rwku} suffers a significant utility loss from $0.5979$ to $0.3495$, but when {\ours} is applied, utility rises back to $0.5253$, nearly matching the original pretrained performance. \underline{Second}, beyond utility preservation, the forget efficacy (FE)—remains either unaffected or slightly \textit{improved} when {\ours} is employed. This balance between utility preservation and effective unlearning highlights the advantage of {\ours}. For instance, \gdiff+{\ours} reduces the FE on Qwen (WMDP) from $0.2964$ to $0.2445$, demonstrating better unlearning while still achieving a higher utility score. Similarly, \rmu+{\ours} on DeepSeek (WMDP) lowers the FE from $0.2530$ to $0.2859$, with a corresponding utility improvement from $0.4540$ to $0.5424$. 
Notably, methods such as {\gdiff} and {\rmu}, which experience notable utility loss when used alone, benefit greatly from the application of {\ours}, achieving near-pretrained utility levels while still maintaining effective unlearning.

\ding{70}  \textbf{{\ours}  outperforms   parameter-efficient fine-tuning (PEFT) methods when used for unlearning.} 
Tab.\, \ref{tab: main_results} also includes a set of baselines that apply  PEFT on GA. It is used to evaluate whether our method unlearns more effectively a subset of parameters (top-1 expert) compared to PEFT.
Tab.\,\ref{tab: peft_param_ratio} shows a comparison of the parameter efficiency involved in tuning.
%\textbf{Tab.\,\ref{tab: peft_results}} shows the performance comparison between {\ours} and other PEFT methods, and Tab.\,\ref{tab: peft_param_ratio} shows a comparison of the parameter efficiency among different PEFT methods. 
The key conclusion from these results is: {\ours} achieves far \textbf{better parameter efficiency}, with only $0.06\%$ of tunable parameters, compared to LoRA ($0.92\%$) and ESFT ($2.86\%$), while still maintaining a comparable level of forget efficacy and \textbf{outperforming them in utility} preservation. For instance, in RWKU, GA+{\ours} achieves utility scores of 0.5709 on Qwen and 0.5485 on DeepSeek, significantly higher than LoRA (0.2689 and 0.2302) and ESFT (0.4433 and 0.5001). %\underline{Second}, the utility preservation of {\ours} is much better than the others, and this is achieved while maintaining a comparable level of forget efficacy. For example, on WMDP, {\ours} achieves a utility score of 0.5012 for Qwen, much higher than LoRA’s 0.2689, with a similar forget efficacy (FE: 0.2987 vs. 0.2459). These results clearly demonstrate that {\ours} is the more balanced and efficient solution for unlearning tasks, particularly when both parameter efficiency and utility retention are important.

\begin{table}\vspace*{-0.2em}
\centering
\caption{\footnotesize  Tunable parameter ratio, PEFT vs \ours. }
%Tunable parameter ratio of the PEFT methods studied in this work for Qwen and DeepSeek.}
\vspace*{-0.5em}
\resizebox{.35\textwidth}{!}{\begin{tabular}{c|ccc}

\toprule[1pt]
\midrule
\multirow{2}{*}{\textbf{Method}} & \multicolumn{3}{c}{\textbf{Tunable Parameter Ratio}}\\
& \textbf{Qwen} & \textbf{DeepSeek} &\textbf{{Mixtral}} \\
\midrule

\textbf{LoRA} &$0.87\%$ & $0.92\%$ &{0.26\%}\\
\textbf{ESFT} & 3.13\% & $2.86\%$&{14\%}\\
\midrule
\textbf{{\ours}} & $0.06\%$ & $0.06\%$&{0.41\%}\\
\midrule
\bottomrule[1pt]
\end{tabular}}
\label{tab: peft_param_ratio}
\vspace*{-1em}
\end{table}
% \begin{wrapfigure}{r}{0.4\linewidth}
%     \vspace*{-1.2em}
%     \centering
%     \includegraphics[width=0.8\linewidth]{figs/overlap_ratio_tune_all_layer.pdf}
%     \vspace*{-.5em}
%     \caption{\footnotesize Expert selection overlap ratio between the pretrained model and the unlearned model across different unlearning iterations using {\ga} on the {\wmdp} benchmark. Experts with the highest affinity score in each layer are tuned during unlearning.}
%     \label{fig: all_layer_overlap_ratio}
%     \vspace*{-1em}
% \end{wrapfigure}

\ding{70} \textbf{Top-1 expert selection outperforms random selection in unlearning}.
In the last row of Tab.~\ref{tab: main_results}, we compare the performance of the affinity score-based expert selection in {\ours} with a random expert selection approach. The results show that while random selection can sometimes preserve utility at a comparable level, it falls short in achieving effective unlearning. For instance, on Qwen ({\wmdp}), random selection yields a higher utility score (0.5947 vs. 0.5351 for {\ours}), but its forget efficacy (FE) remains significantly higher (0.3505 vs. 0.2536 for {\ours}), indicating incomplete unlearning. This suggests that selecting the top-1 expert based on affinity scores is crucial for reducing FE while maintaining utility, making it a superior approach to random selection.

\ding{70} \textbf{Experts with higher affinity scores play a more significant role in  unlearning}.
To further examine the impact of selecting experts based on their affinity scores, we analyze the layer-wise Top-1 expert in DeepSeek on RWKU dataset. In Tab.\,\ref{tab: other_layer_selection}, we present their affinity scores along with the utility (UT) when the expert is involved in unlearning. Due to space constraints, we highlight the top-ranked layer-wise experts (1st to 3rd) and also include several lower-ranked ones (13th to 26th) for comparison. 
From the results, we observe that the first-ranked expert (with the highest affinity score 0.211) yields the highest UT (0.5485). Overall, UT remains stable at 0.5445 or higher when selecting experts with affinity scores above 0.1. However, when affinity scores drop further (e.g., the 23rd and 26th ranked experts), utility declines more sharply to 0.4262 and 0.2355. These findings emphasize the importance of selecting experts with sufficiently high affinity scores to maintain utility while achieving effective unlearning.

\begin{table}
    % \vspace*{-.2em}
    \centering
    \caption{\footnotesize Model utility (UT) comparison across 
    unlearned experts with different affinity scores ($s_i$)  in {\ours}+RMU on the {\rwku} benchmark. UT is compared at a consistent level of forget efficacy (FE $\approx 0.25$).}
    % \vspace*{-.5em}
    \resizebox{\linewidth}{!}{
    \begin{tabular}{c|ccccccc}
    \toprule[1pt]
    \midrule
     Rank & \#1 &{\#2}&{\#3} & \#13 & \#20 & \#23 & \#26 \\  \midrule
    $s_i$ & 0.2110&{0.1957}&{0.1695} & 0.1115 & 0.0942 & 0.0844 & 0.0618 \\
     \midrule

     UT ($\uparrow$) & 0.5485 &{0.5475}&{0.5453}& 0.5445 & 0.5441 & 0.4262 & 0.2355 \\
     \midrule
    \bottomrule[1pt]
    \end{tabular}
    }
    \label{tab: other_layer_selection}
    \vspace*{-.5em}
\end{table}

%$\bullet$ \textit{Sensitivity of {\ours} to layer selection schemes.} We examine the performance of {\ours} when layers with different top1 expert affinity score rankings are selected for unlearning. In Tab.\,\ref{tab: other_layer_selection}, we observe that {\ours} is robust and relatively insensitive to the specific layer chosen for unlearning, as long as the affinity score remains reasonably high. For instance, even when selecting the 13th or 20th ranked layers, the model utility (\textbf{UT}) remains stable at around 0.5445, although their  affinity scores of 0.1115 and 0.0942 are lower than that of the top-ranked layer. However, once the affinity score drops further, as seen in the 23rd and 26th ranked layers (with scores of 0.0844 and 0.0618), the utility decreases more sharply, falling to 0.4262 and 0.2355, respectively. This demonstrates that while {\ours} maintains strong performance across a wide range of layers, selecting layers with very low affinity scores can negatively impact utility. Overall, these results highlight the robustness of {\ours} and its ability to tolerate variability in layer selection without sacrificing unlearning efficacy, provided that layers with sufficiently high affinity scores are chosen.

\begin{figure}[t]
    \centering
    
    \includegraphics[width=\linewidth]{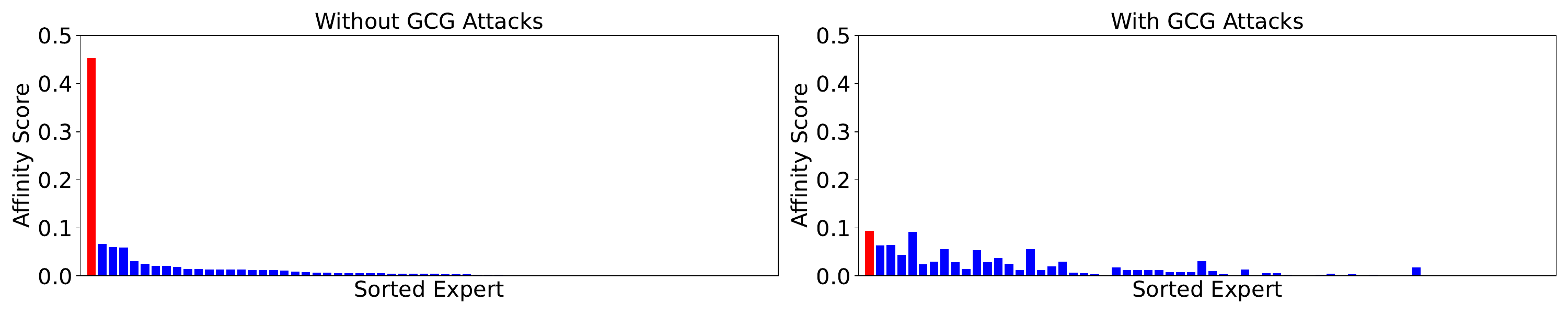} \vspace{-.3in}
    \caption{{\footnotesize Comparison of affinity scores for all experts in the target layer of DeepSeek unlearned by {\ours} + GA on the RWKU dataset, with and without the GCG attack. The target expert is marked as red.}}
    \vspace{-.2in}\label{fig:visual_gcg_attack_expert_selection}
\end{figure}

\ding{70} \textbf{Unlearning resilient to jailbreak attacks}. The unlearned model is expected to refuse harmful queries. The  forgotten knowledge  should  not be recovered even through adversarial means. We thus examine the behavior of  MoE LLMs unlearned by {\ours}  under adversarial prompting. Specifically, we test whether {\ours} effectively mitigates unauthorized responses by employing the Greedy Coordinate Gradient (GCG) attack \cite{zou2023universal} in a white-box setting. This attack optimizes attack prompts to elicit responses that begin with ``Sure, here is the answer:''.  To increase attack strength, we extend the number of optimization steps to 5,000, while keeping other hyperparameters at their default settings. Given the computational cost ($\sim$ 1 GPU hour on an A100 per soft prompt), we optimize 400 prompts across 400 samples in RWKU for attacking DeepSeek unlearned by {\ours}+GA. Since not all responses explicitly begin with "Sure, here is the answer:", we filter for outputs containing the word "answer" and evaluate forget efficacy (FE) both with and without GCG-generated prompts.
Our results show that despite being one of the strongest prompt-level attacks, GCG fails to recover forgotten knowledge, as \textbf{FE remains at 0.01 before and after the attack}.
To further understand how the GCG attack affects expert selection, we visualize the affinity score of experts in DeepSeek, and compare it with GCG-attacked DeepSeek.
 Fig.~\ref{fig:visual_gcg_attack_expert_selection} shows that while the GCG attack reduces the affinity score of the target expert, the expert remains ranked as the top-1 in affinity score. This suggests that {\ours} maintains stable expert selection even under adversarial influence, ensuring robustness in the unlearning process.

%\textbf{Additional experiments.} To assess the robustness of {\ours} against Jailbreak attacks, we apply the gradient-based Jailbreak attack GCG~\cite{zou2023universal} to \ours. As shown in Sec.~\ref{sec: jailbreak} in Appendix, {\ours} remains robust to Jailbreak attacks, despite shifts in expert selection. 
Additionally, we also perform a sensitivity analysis on hyperparameter $\alpha$ in Sec.~\ref{sec:hyperparameter} in Appendix. The results in Tab.~\ref{table: Sensitivity} in Appendix indicate that $\alpha = 1$ achieves the best performance.

\vspace{-0.2cm}
\section{Conclusion}
\vspace{-0.2cm}
In this paper, we for the first time examine the challenges of applying existing MU techniques to MoE LLMs and carefully investigate the synergy between the dynamic routing system of MoE LLM and the unlearning effects. To address these issues, we proposed {\ours}, a novel framework that unlearns most related experts while stabilizing expert selection through a router anchor loss. This approach mitigates expert selection shifts and achieves efficient unlearning with minimal parameter updates. Extensive experiments show that {\ours} significantly outperforms traditional unlearning methods and other parameter-efficient fine-tuning techniques, providing a robust solution for MoE LLM unlearning tasks.

\section{Limitation}
While this study offers valuable insights into unlearning of MoE LLMs, it has certain limitations. First, the evaluation was limited to two datasets due to the scarcity of standardized benchmarks in unlearning. We have used two widely accepted LLM unlearning benchmarks: {\wmdp} \citep{li2024wmdp} and {\rwku} \citep{jin2024rwku}. {\wmdp}.   We acknowledge the existence of other commonly used benchmarks, such as TOFU \citep{maini2024tofu} and MUSE \citep{shi2024muse}. However, these benchmarks are \textbf{less suitable} for our study, as they require models to \textbf{undergo fine-tuning before unlearning}. This additional training step introduces biases in MoE LLMs due to known instability in training, sensitive hyperparameter tuning, and the risk of training collapse \citep{jiang2024mixtral, zoph2022designing}. These factors make it challenging to isolate the effects of unlearning from the broader impact of model fine-tuning.
Expanding the evaluation to a broader range of datasets could enhance the generalizability of the findings. 
In future work, we plan to explore additional benchmarks, including those that do not require fine-tuning before unlearning, to ensure a more comprehensive assessment of unlearning effectiveness across diverse tasks and model architectures.
Second, the study did not apply the unlearning algorithm to Mixtral 8×7B with all parameters unlearned and excluded larger MoE LLM models like DeepSeek-R1 due to computational constraints. Due to the computation limitation, Mixtral is only applied on {\ours} and other parameter-efficient fine-tuning unlearning baselines. In future work, we could explore scaling the approach to larger models to evaluate its effectiveness in more complex architectures.

\section{Acknowledgement}
This work is supported by the NSF award \# 2321054.

% \clearpage

% \bibliographystyle{iclr2025_conference}
\clearpage
\appendix

% \begin{figure*}[h]
%     \centering
    
%     \includegraphics[width=0.9\linewidth]{figs/layer_example_gating_selection.pdf}
%     \caption{{\footnotesize Token Proportion Shift in layer 21 of Deepseek unlearned by GA in WMDP dataset.}}
%     \label{fig:enter-label}
% \end{figure*}

\iffalse
\begin{table*}[t!]
    \centering
    \caption{{Example of GCG attack. Original is the output of the original Deepseek model.}}
    \resizebox{0.7\linewidth}{!}{
    \begin{tabular}{p{3cm}|p{1.5cm}|p{3cm}|p{3cm}}
    \toprule[1.5pt]
        Questions&Original & Without Attack&With Attack \\ \midrule
        Stephen King is often referred to as the 'King of \_\_\_'. &Horror & indingindingind-ingindingindingi-ndinginding…&ologist:Sure here is the answer:
Question : Stephenindingindingin-dinginding…\\ \midrule
Bruce Lee moved to \_\_\_ in 1959. &America&-wardswar-dswardswards...&eanswer:"
Answer:[/b]inding [b]...
\\ \bottomrule[1.5pt]
    \end{tabular}}
    
    \label{tab:GCG_attack_example}
\end{table*}
\fi
\section{{Experiment Setups}}
\label{Sec: setup}
\textbf{Unlearning tasks and datasets.}  To demonstrate the effectiveness of our proposed method, we evaluate and compare it against different baselines on two widely accepted LLM unlearning benchmarks: {\wmdp} \citep{li2024wmdp} and {\rwku} \citep{jin2024rwku}. {\wmdp} assesses the model’s ability to unlearn and prevent the generation of hazardous knowledge in biosecurity, cybersecurity, and chemical security contexts. {\rwku}, on the other hand, evaluates the model’s capability to eliminate knowledge about 200 real-world celebrities, simulating a private information protection task. We follow the original study by selecting 100 individuals as unlearning targets. The train\_original\_passage set, which includes Wikipedia descriptions of these 100 individuals as provided in the paper, is used as the forget set.
We note that other commonly used benchmarks, such as TOFU \citep{maini2024tofu} and MUSE \citep{shi2024muse}, are less appropriate in this work. These benchmarks require models to be fine-tuned before unlearning, which introduces additional biases to the results for MoE LLMs due to the known instability in training and the tricky hyper-parameter tuning involved \citep{jiang2024mixtral}, often leading to training collapse \citep{zoph2022designing}.

\textbf{Target MoE models to unlearn.} We evaluate different unlearning methods on two MoE LLMs: Qwen1.5-MoE-A2.7B-Chat (Qwen), {mistralai/Mixtral-8x7B-Instruct-v0.1 (Mixtral)}, and DeepSeek-V2-Lite (DeepSeek), representing the two mainstream MoE LLM training schemes: upcycle-from-dense and train-from-scratch, respectively. Qwen has a total of 14.3 billion parameters, with 2.7 billion activated during inference, while DeepSeek has 16 billion parameters, of which 2.4 billion are activated during inference. Mixtral has 45 billion parameters, of which 12.9 billion are activated. 

\textbf{Evaluation setup.} We evaluate the performance of the unlearned LLMs based on two key metrics: forget efficacy (\textbf{FE}) and preserved model utility (\textbf{UT}). For the \textbf{{\wmdp}} task, FE is measured using the {\wmdp}-Cyber subsets provided by the benchmark. Specifically, we use the accuracy of the forget set after unlearning as the measure of FE. A lower FE indicates better unlearning.  Given the four-option multiple-choice format of the test set, the ideal FE is $0.25$, equivalent to random guessing. UT is assessed using the zero-shot accuracy on the MMLU dataset \citep{hendrycks2020measuring}, which reflects the model’s ability to retain general knowledge.
For the \textbf{{\rwku}} task, we use the Rouge-L recall score to evaluate performance on fill-in-the-blank and question-answer tasks, with lower scores indicating more effective unlearning. Since the task follows a question-answer format, the ideal FE is $0.0$, indicating no overlap between the generated answer and the ground truth. The UT evaluation for {\rwku} is the same as for {\wmdp}, using the MMLU benchmark. By default, during the unlearning process, we select the model checkpoint that achieves the best balance between FE and UT as the optimal checkpoint.

We utilize the LM Evaluation Harness~\citep{eval-harness} to measure zero-shot accuracy on the MMLU and WMDP cyber datasets. The mean accuracy across all tasks in MMLU serves as a measure of model utility. For the RWKU dataset, we adhere to the original settings, using the prompt ``Please complete the blank in the following question. Question:" for fill-in-the-blank tasks and ``Please briefly answer the following question. Question:" for generation tasks.

\textbf{Unlearning Baselines.} We demonstrate the effectiveness of our proposed {\ours} framework by comparing it against the LLM unlearning baselines: Gradient Ascent ({\ga})~\citep{eldan2023whos}, Gradient Difference ({\gdiff})~\citep{maini2024tofu} and most recent unlearning algorithm Negative Preference Optimization ({\npo})~\citep{zhang2024negative} and Representation Misdirection for Unlearning ({\rmu})~\citep{li2024wmdp}. For each method, we compare the original results with those obtained when incorporating {\ours}. Given the parameter efficiency of {\ours}, we also compare it with two state-of-the-art parameter-efficient fine-tuning (PEFT) methods for MoE LLMs: the low-rank adaptation scheme (LoRA) \citep{hu2021lora} and the Expert-Specialized Fine-Tuning method (ESFT) \cite{wang2024let}, which is specifically designed for MoE LLMs.

\textbf{Hyperparameter selection.} We consider typical unlearning algorithm as baselines. For {\rmu}, due to the original parameters settings for MoE models fail to unlearn both in DeepSeek and Qwen. We adapt its settings to target all expert MLP layers in fifth, sixth, seventh layers, which align with the settings in the dense model. For the hyperparameters, the retain effect parameter is set to 1200, and $c$ is set to 30000 and 3000 in DeepSeek and Qwen, respectively. We set the learning rate to 5e-5 for GA, NPO, and GD while setting it to 1e-4 for \ours. The batch size is 4 for GA, NPO, and GD, while it is set to 16 for \ours. In NPO, the beta value is set to 0.001. The $\lambda$ for the retain loss is set to 1 in both GD and NPO. For RMU, we follow the hyperparameters specified in the original work. We configure the steering coefficients as 8000 for Qwen and 32000 for Deepseek, as {\ours} targets deeper layers in these models. For ESFT, we set the threshold $p=0.15$. According to Insight 4 in Sec.~\ref{sec: method}, we set $M=1$ in the experiment section by default. For LoRA, we apply low-rank adaptation to all layers of the model to enable full-layer fine-tuning. All experiments were conducted in a single run without multiple trials.

\section{Sensitivity Analysis of  $\alpha$}
\label{sec:hyperparameter}
The hyperparameter $\alpha$ is used for ancho loss in our loss function 
\(
 \min_{\btheta} \ell_f(\btheta; \mathcal{D}_f) + \lambda \ell_r(\btheta; \mathcal{D}_r)+\alpha L_{\text{anchor}}^{(l)},\) for introducing the anchor loss $L_{\text{anchor}}$.
{We conduct experiments on Deepseek unlearned by {\ga} with RWKU dataset to explore the performance of different $\alpha$. As shown in Tab.~\ref{table: Sensitivity}, the results indicate that {\ours} is robust to a wide range of $\alpha$ and achieves the best performance when $\alpha=1$.}
\begin{table}[h]
\centering
\caption{{Sensitivity Analysis of hyperparameter $\alpha$ for the strength of anchor loss. The experiment is conducted on Deepseek unlearned by {\ga} with RWKU dataset.} }
\resizebox{0.8\linewidth}{!}{
\begin{tabular}{c|cccc}
\toprule[1.5pt]
$\alpha$ &0 & 1      & 100    & 1000   \\ \midrule
FE ($\downarrow$) &0.0& 0.0    & 0.0    & 0.0       \\
UT ($\uparrow$) &0.5435& 0.5485 & 0.5471 & 0.5468   \\
\bottomrule[1.5pt]
\end{tabular}}

\label{table: Sensitivity}
\end{table}

\section{Selection of top $M$ experts in different tasks}
We also conduct experiments on Qwen unlearned by {\ga} with {\rwku} dataset to investigate the optimal selection of $M$. The results in Tab.~\ref{tab:NumExpert_qwen_rwku} indicate that {\ours} achieves the best performance when only one expert is unlearned $M=1$, which is consistent with the Insight 4.
\begin{table}[h]
\centering
\caption{\footnotesize Model utility (UT$\uparrow$) comparison at the same level of forget efficacy (FE$\approx0.25$),
    when the top $M$ experts from either the same layer or different layers in Qwen are unlearned using {\ga} on {\rwku} benchmark, also when  4 shared experts are included.  } \vspace{-0.1in}
\label{tab:NumExpert_qwen_rwku}

\scalebox{0.72}{\begin{tabular}{l|llcc}
\toprule
\multicolumn{1}{l|}{Selected experts}                       & \multicolumn{1}{c}{Top-1}      & \multicolumn{1}{c}{Top-3}      & Top-6   & Top-1+4-shared   \\ \midrule

\multicolumn{1}{c|}{Same layer}                           &   \multicolumn{1}{l}{0.5709} & \multicolumn{1}{l}{0.3695} & 0.2572 & 0.2445 \\ \midrule

\multicolumn{1}{l|}{Different layers}                           &  \multicolumn{1}{l}{0.5709} & \multicolumn{1}{l}{0.4224} & 0.3872  & -\\ \bottomrule
\end{tabular}}
\vspace{-0.2in}

\end{table}

\section{Robustness of Expert Selection}
To evaluate the robustness of expert selection under token sampling, we conducted an additional experiment on a consistency analysis on the DeepSeek-V2-Lite model using the WMDP forget set. Specifically, we computed the overlap ratio of selected experts across different token subsets, where overlap is defined as the proportion of shared top-6 experts at each MoE layer.

As shown in Table~\ref{tab: overlap}, a subset of 100,000 tokens yields a high overlap (0.94) with the expert selections derived from the full dataset. Furthermore, two independently sampled subsets also show strong agreement with each other (0.87 overlap), indicating that the attribution process is stable across different sampling runs.
\begin{table}[t]
\centering
\caption{Expert selection overlap between different sampling splits}
\label{tab: overlap}
\begin{tabular}{lccc}
\toprule
\textbf{Dataset} & \textbf{Full} & \textbf{Subset 1} & \textbf{Subset 2} \\
\midrule
Full       & 1.00 & 0.94 & 0.85 \\
Subset 1   & 0.94 & 1.00 & 0.87 \\
Subset 2   & 0.85 & 0.87 & 1.00 \\
\bottomrule
\end{tabular}
\end{table}

\section{Experiments on Larger MoE Models}

\begin{table}[t]
\centering
\caption{Performance of Mixtral 8x7B unlearned by GA on WMDP and RWKU datasets.}
\label{tab: Mixtral}
\begin{tabular}{lcccc}
\toprule
\textbf{Method} & \multicolumn{2}{c}{\textbf{WMDP}} & \multicolumn{2}{c}{\textbf{RWKU}} \\
 & \textbf{FE ↓} & \textbf{UT ↑} & \textbf{FE ↓} & \textbf{UT ↑} \\
\midrule
Pretrained & 0.5229 & 0.6885 & 0.5820 & 0.6885 \\
LoRA       & 0.2658 & 0.2597 & 0.0000 & 0.2295 \\
ESFT       & 0.2574 & 0.6386 & 0.0542 & 0.6743 \\
SEUF       & 0.2608 & 0.6364 & 0.0455 & 0.6713 \\
\bottomrule
\end{tabular}
\end{table}

\begin{table}[t]
\centering
\caption{Tunable parameter ratio of different methods}
\label{tab: parameter_ratio}
\begin{tabular}{lc}
\toprule
\textbf{Method} & \textbf{Tunable Parameter Ratio $\downarrow$} \\
\midrule
LoRA & 0.26\% \\
ESFT & 14\% \\
SEUF & 0.41\% \\
\bottomrule
\end{tabular}
\end{table}

To explore if {\ours} can be applied to larger MoE models, we evaluated SEUF on mistralai/Mixtral-8x7B-Instruct-v0.1 (Mixtral 8x7B), one of the most widely used large-scale open-source MoE models, and compared its performance to other parameter-efficient unlearning baselines.

As shown in Table~\ref{tab: Mixtral}, SEUF achieves comparable or even better utility (UT) while maintaining strong forget efficacy (FE). On the WMDP dataset, SEUF achieves a UT of 0.6364, close to ESFT’s 0.6386 and far better than LoRA’s 0.2597. On RWKU, SEUF reaches 0.6713, again comparable to ESFT (0.6743) and significantly ahead of LoRA (0.2295). Importantly, SEUF does so while updating only 0.41\% of parameters, as shown in Table~\ref{tab: parameter_ratio}, substantially fewer than ESFT’s 14\%.

\section{The Affect of Weighted Expert Norms}

\begin{table}[t]
\centering
\caption{Table A: The Spearman’s rank correlation between $g_{i,t}$ and $g_{i,t} E(x_i)$ for All experts and Top 6 experts across all layers in DeepSeek.}
\label{tab: correlation}
\begin{tabular}{lcc}
\toprule
\textbf{Range} & \textbf{All experts} & \textbf{Top 6} \\
\midrule
Correlation & 1.0 & 1.0 \\
\bottomrule
\end{tabular}
\end{table}
\vspace{0.2cm}

As experts may have different weight norms, which could in theory impact the total contribution of their outputs, and that $g_{i,t} E(x_i)$ might reflect this better than $g_{i,t}$ alone. To investigate this, we computed the Spearman’s rank correlation between $g_{i,t}$ and $g_{i,t} E(x_i)$ across all MoE layers in DeepSeek using the WMDP dataset.
As shown in the Tab~\ref{tab: correlation}, the average rank correlation is 1.0 across both all experts and the top-6 experts, indicating that the ordering induced by $g_{i,t}$ closely matches that of $g_{i,t} E(x_i)$. This suggests that gating scores alone already serve as a strong proxy for expert contribution, even without explicitly incorporating the output norms. This result aligns with the design of the MoE architecture, where routing is learned independently per token while expert weights are optimized to produce scale-compatible outputs under the sparse gating mechanism.


\begin{thebibliography}{69}
\providecommand{\natexlab}[1]{#1}

\bibitem[{Abdin et~al.(2024)Abdin, Jacobs, Awan, Aneja, Awadallah, Awadalla, Bach, Bahree, Bakhtiari, Behl et~al.}]{abdin2024phi}
Marah Abdin, Sam~Ade Jacobs, Ammar~Ahmad Awan, Jyoti Aneja, Ahmed Awadallah, Hany Awadalla, Nguyen Bach, Amit Bahree, Arash Bakhtiari, Harkirat Behl, et~al. 2024.
\newblock Phi-3 technical report: A highly capable language model locally on your phone.
\newblock \emph{arXiv preprint arXiv:2404.14219}.

\bibitem[{Barbulescu and Triantafillou(2024)}]{barbulescu2024each}
George-Octavian Barbulescu and Peter Triantafillou. 2024.
\newblock To each (textual sequence) its own: Improving memorized-data unlearning in large language models.
\newblock \emph{arXiv preprint arXiv:2405.03097}.

\bibitem[{Chowdhery et~al.(2023)Chowdhery, Narang, Devlin, Bosma, Mishra, Roberts, Barham, Chung, Sutton, Gehrmann et~al.}]{chowdhery2023palm}
Aakanksha Chowdhery, Sharan Narang, Jacob Devlin, Maarten Bosma, Gaurav Mishra, Adam Roberts, Paul Barham, Hyung~Won Chung, Charles Sutton, Sebastian Gehrmann, et~al. 2023.
\newblock Palm: Scaling language modeling with pathways.
\newblock \emph{Journal of Machine Learning Research}, 24(240):1--113.

\bibitem[{Cong et~al.(2024)Cong, Yuan, Chen, Tian, Ye, and Yang}]{cong2024prediction}
Peizhuang Cong, Aomufei Yuan, Shimao Chen, Yuxuan Tian, Bowen Ye, and Tong Yang. 2024.
\newblock Prediction is all moe needs: Expert load distribution goes from fluctuating to stabilizing.
\newblock \emph{arXiv preprint arXiv:2404.16914}.

\bibitem[{Dai et~al.(2024)Dai, Deng, Zhao, Xu, Gao, Chen, Li, Zeng, Yu, Wu et~al.}]{dai2024deepseekmoe}
Damai Dai, Chengqi Deng, Chenggang Zhao, RX~Xu, Huazuo Gao, Deli Chen, Jiashi Li, Wangding Zeng, Xingkai Yu, Y~Wu, et~al. 2024.
\newblock Deepseekmoe: Towards ultimate expert specialization in mixture-of-experts language models.
\newblock \emph{arXiv preprint arXiv:2401.06066}.

\bibitem[{Dai et~al.(2022)Dai, Dong, Ma, Zheng, Sui, Chang, and Wei}]{dai2022stablemoe}
Damai Dai, Li~Dong, Shuming Ma, Bo~Zheng, Zhifang Sui, Baobao Chang, and Furu Wei. 2022.
\newblock Stablemoe: Stable routing strategy for mixture of experts.
\newblock \emph{arXiv preprint arXiv:2204.08396}.

\bibitem[{Databricks(2024)}]{databricks_dbrx_2024}
Databricks. 2024.
\newblock Introducing dbrx: A new state-of-the-art open llm.
\newblock \url{https://www.databricks.com/blog/introducing-dbrx-new-state-art-open-llm}.
\newblock Accessed: 2024-09-25.

\bibitem[{Dubey et~al.(2024)Dubey, Jauhri, Pandey, Kadian, Al-Dahle, Letman, Mathur, Schelten, Yang, Fan et~al.}]{dubey2024llama}
Abhimanyu Dubey, Abhinav Jauhri, Abhinav Pandey, Abhishek Kadian, Ahmad Al-Dahle, Aiesha Letman, Akhil Mathur, Alan Schelten, Amy Yang, Angela Fan, et~al. 2024.
\newblock The llama 3 herd of models.
\newblock \emph{arXiv preprint arXiv:2407.21783}.

\bibitem[{Eldan and Russinovich(2023)}]{eldan2023whos}
Ronen Eldan and Mark Russinovich. 2023.
\newblock \href {https://arxiv.org/abs/2310.02238} {Who's harry potter? approximate unlearning in llms}.
\newblock \emph{Preprint}, arXiv:2310.02238.

\bibitem[{Fedus et~al.(2022)Fedus, Zoph, and Shazeer}]{fedus2022switch}
William Fedus, Barret Zoph, and Noam Shazeer. 2022.
\newblock Switch transformers: Scaling to trillion parameter models with simple and efficient sparsity.
\newblock \emph{Journal of Machine Learning Research}, 23(120):1--39.

\bibitem[{Gale et~al.(2023)Gale, Narayanan, Young, and Zaharia}]{gale2023megablocks}
Trevor Gale, Deepak Narayanan, Cliff Young, and Matei Zaharia. 2023.
\newblock Megablocks: Efficient sparse training with mixture-of-experts.
\newblock \emph{Proceedings of Machine Learning and Systems}, 5:288--304.

\bibitem[{Gao et~al.(2024)Gao, Tow, Abbasi, Biderman, Black, DiPofi, Foster, Golding, Hsu, Le~Noac'h, Li, McDonell, Muennighoff, Ociepa, Phang, Reynolds, Schoelkopf, Skowron, Sutawika, Tang, Thite, Wang, Wang, and Zou}]{eval-harness}
Leo Gao, Jonathan Tow, Baber Abbasi, Stella Biderman, Sid Black, Anthony DiPofi, Charles Foster, Laurence Golding, Jeffrey Hsu, Alain Le~Noac'h, Haonan Li, Kyle McDonell, Niklas Muennighoff, Chris Ociepa, Jason Phang, Laria Reynolds, Hailey Schoelkopf, Aviya Skowron, Lintang Sutawika, Eric Tang, Anish Thite, Ben Wang, Kevin Wang, and Andy Zou. 2024.
\newblock \href {https://doi.org/10.5281/zenodo.12608602} {A framework for few-shot language model evaluation}.

\bibitem[{Hendrycks et~al.(2020)Hendrycks, Burns, Basart, Zou, Mazeika, Song, and Steinhardt}]{hendrycks2020measuring}
Dan Hendrycks, Collin Burns, Steven Basart, Andy Zou, Mantas Mazeika, Dawn Song, and Jacob Steinhardt. 2020.
\newblock Measuring massive multitask language understanding.
\newblock \emph{arXiv preprint arXiv:2009.03300}.

\bibitem[{Hendrycks et~al.(2023)Hendrycks, Mazeika, and Woodside}]{hendrycks2023overview}
Dan Hendrycks, Mantas Mazeika, and Thomas Woodside. 2023.
\newblock An overview of catastrophic ai risks.
\newblock \emph{arXiv preprint arXiv:2306.12001}.

\bibitem[{Hong et~al.(2024)Hong, Lee, and Thorne}]{hong2024reference}
Jiwoo Hong, Noah Lee, and James Thorne. 2024.
\newblock Reference-free monolithic preference optimization with odds ratio.
\newblock \emph{arXiv preprint arXiv:2403.07691}.

\bibitem[{Hu et~al.(2021)Hu, Shen, Wallis, Allen-Zhu, Li, Wang, Wang, and Chen}]{hu2021lora}
Edward~J Hu, Yelong Shen, Phillip Wallis, Zeyuan Allen-Zhu, Yuanzhi Li, Shean Wang, Lu~Wang, and Weizhu Chen. 2021.
\newblock Lora: Low-rank adaptation of large language models.
\newblock \emph{arXiv preprint arXiv:2106.09685}.

\bibitem[{Hu et~al.(2024)Hu, Li, Hu, Zheng, Liu, and Zhang}]{hu2024separate}
Xinshuo Hu, Dongfang Li, Baotian Hu, Zihao Zheng, Zhenyu Liu, and Min Zhang. 2024.
\newblock Separate the wheat from the chaff: Model deficiency unlearning via parameter-efficient module operation.
\newblock In \emph{Proceedings of the AAAI Conference on Artificial Intelligence}, volume~38, pages 18252--18260.

\bibitem[{Hwang et~al.(2023)Hwang, Cui, Xiong, Yang, Liu, Hu, Wang, Salas, Jose, Ram et~al.}]{hwang2023tutel}
Changho Hwang, Wei Cui, Yifan Xiong, Ziyue Yang, Ze~Liu, Han Hu, Zilong Wang, Rafael Salas, Jithin Jose, Prabhat Ram, et~al. 2023.
\newblock Tutel: Adaptive mixture-of-experts at scale.
\newblock \emph{Proceedings of Machine Learning and Systems}, 5:269--287.

\bibitem[{Ilharco et~al.(2022)Ilharco, Ribeiro, Wortsman, Gururangan, Schmidt, Hajishirzi, and Farhadi}]{ilharco2022editing}
Gabriel Ilharco, Marco~Tulio Ribeiro, Mitchell Wortsman, Suchin Gururangan, Ludwig Schmidt, Hannaneh Hajishirzi, and Ali Farhadi. 2022.
\newblock Editing models with task arithmetic.
\newblock \emph{arXiv preprint arXiv:2212.04089}.

\bibitem[{Ishibashi and Shimodaira(2023)}]{ishibashi2023knowledge}
Yoichi Ishibashi and Hidetoshi Shimodaira. 2023.
\newblock Knowledge sanitization of large language models.
\newblock \emph{arXiv preprint arXiv:2309.11852}.

\bibitem[{Jang et~al.(2022)Jang, Yoon, Yang, Cha, Lee, Logeswaran, and Seo}]{jang2022knowledge}
Joel Jang, Dongkeun Yoon, Sohee Yang, Sungmin Cha, Moontae Lee, Lajanugen Logeswaran, and Minjoon Seo. 2022.
\newblock Knowledge unlearning for mitigating privacy risks in language models.
\newblock \emph{arXiv preprint arXiv:2210.01504}.

\bibitem[{Jia et~al.(2024{\natexlab{a}})Jia, Liu, Zhang, Ram, Baracaldo, and Liu}]{jia2024wagle}
Jinghan Jia, Jiancheng Liu, Yihua Zhang, Parikshit Ram, Nathalie Baracaldo, and Sijia Liu. 2024{\natexlab{a}}.
\newblock Wagle: Strategic weight attribution for effective and modular unlearning in large language models.
\newblock \emph{arXiv preprint arXiv:2410.17509}.

\bibitem[{Jia et~al.(2024{\natexlab{b}})Jia, Zhang, Zhang, Liu, Runwal, Diffenderfer, Kailkhura, and Liu}]{jia2024soul}
Jinghan Jia, Yihua Zhang, Yimeng Zhang, Jiancheng Liu, Bharat Runwal, James Diffenderfer, Bhavya Kailkhura, and Sijia Liu. 2024{\natexlab{b}}.
\newblock Soul: Unlocking the power of second-order optimization for llm unlearning.
\newblock \emph{arXiv preprint arXiv:2404.18239}.

\bibitem[{Jiang et~al.(2024)Jiang, Sablayrolles, Roux, Mensch, Savary, Bamford, Chaplot, Casas, Hanna, Bressand et~al.}]{jiang2024mixtral}
Albert~Q Jiang, Alexandre Sablayrolles, Antoine Roux, Arthur Mensch, Blanche Savary, Chris Bamford, Devendra~Singh Chaplot, Diego de~las Casas, Emma~Bou Hanna, Florian Bressand, et~al. 2024.
\newblock Mixtral of experts.
\newblock \emph{arXiv preprint arXiv:2401.04088}.

\bibitem[{Jin et~al.(2024)Jin, Cao, Wang, He, Yuan, Li, Chen, Liu, and Zhao}]{jin2024rwku}
Zhuoran Jin, Pengfei Cao, Chenhao Wang, Zhitao He, Hongbang Yuan, Jiachun Li, Yubo Chen, Kang Liu, and Jun Zhao. 2024.
\newblock Rwku: Benchmarking real-world knowledge unlearning for large language models.
\newblock \emph{arXiv preprint arXiv:2406.10890}.

\bibitem[{Kim et~al.(2021)Kim, Awan, Muzio, Salinas, Lu, Hendy, Rajbhandari, He, and Awadalla}]{kim2021scalable}
Young~Jin Kim, Ammar~Ahmad Awan, Alexandre Muzio, Andres Felipe~Cruz Salinas, Liyang Lu, Amr Hendy, Samyam Rajbhandari, Yuxiong He, and Hany~Hassan Awadalla. 2021.
\newblock Scalable and efficient moe training for multitask multilingual models.
\newblock \emph{arXiv preprint arXiv:2109.10465}.

\bibitem[{Komatsuzaki et~al.(2022)Komatsuzaki, Puigcerver, Lee-Thorp, Ruiz, Mustafa, Ainslie, Tay, Dehghani, and Houlsby}]{komatsuzaki2022sparse}
Aran Komatsuzaki, Joan Puigcerver, James Lee-Thorp, Carlos~Riquelme Ruiz, Basil Mustafa, Joshua Ainslie, Yi~Tay, Mostafa Dehghani, and Neil Houlsby. 2022.
\newblock Sparse upcycling: Training mixture-of-experts from dense checkpoints.
\newblock \emph{arXiv preprint arXiv:2212.05055}.

\bibitem[{Kumar et~al.(2022)Kumar, Gangadharaiah, and Roth}]{kumar2022privacy}
Vinayshekhar~Bannihatti Kumar, Rashmi Gangadharaiah, and Dan Roth. 2022.
\newblock Privacy adhering machine un-learning in nlp.
\newblock \emph{arXiv preprint arXiv:2212.09573}.

\bibitem[{Lepikhin et~al.(2020)Lepikhin, Lee, Xu, Chen, Firat, Huang, Krikun, Shazeer, and Chen}]{lepikhin2020gshard}
Dmitry Lepikhin, HyoukJoong Lee, Yuanzhong Xu, Dehao Chen, Orhan Firat, Yanping Huang, Maxim Krikun, Noam Shazeer, and Zhifeng Chen. 2020.
\newblock Gshard: Scaling giant models with conditional computation and automatic sharding.
\newblock \emph{arXiv preprint arXiv:2006.16668}.

\bibitem[{Li et~al.(2024)Li, Pan, Gopal, Yue, Berrios, Gatti, Li, Dombrowski, Goel, Phan et~al.}]{li2024wmdp}
Nathaniel Li, Alexander Pan, Anjali Gopal, Summer Yue, Daniel Berrios, Alice Gatti, Justin~D Li, Ann-Kathrin Dombrowski, Shashwat Goel, Long Phan, et~al. 2024.
\newblock The wmdp benchmark: Measuring and reducing malicious use with unlearning.
\newblock \emph{arXiv preprint arXiv:2403.03218}.

\bibitem[{Lieber et~al.(2024)Lieber, Lenz, Bata, Cohen, Osin, Dalmedigos, Safahi, Meirom, Belinkov, Shalev-Shwartz et~al.}]{lieber2024jamba}
Opher Lieber, Barak Lenz, Hofit Bata, Gal Cohen, Jhonathan Osin, Itay Dalmedigos, Erez Safahi, Shaked Meirom, Yonatan Belinkov, Shai Shalev-Shwartz, et~al. 2024.
\newblock Jamba: A hybrid transformer-mamba language model.
\newblock \emph{arXiv preprint arXiv:2403.19887}.

\bibitem[{Liu et~al.(2024{\natexlab{a}})Liu, Feng, Wang, Wang, Liu, Zhao, Dengr, Ruan, Dai, Guo et~al.}]{liu2024deepseek}
Aixin Liu, Bei Feng, Bin Wang, Bingxuan Wang, Bo~Liu, Chenggang Zhao, Chengqi Dengr, Chong Ruan, Damai Dai, Daya Guo, et~al. 2024{\natexlab{a}}.
\newblock Deepseek-v2: A strong, economical, and efficient mixture-of-experts language model.
\newblock \emph{arXiv preprint arXiv:2405.04434}.

\bibitem[{Liu et~al.(2022)Liu, Liu, and Stone}]{liu2022continual}
Bo~Liu, Qiang Liu, and Peter Stone. 2022.
\newblock Continual learning and private unlearning.
\newblock In \emph{Conference on Lifelong Learning Agents}, pages 243--254. PMLR.

\bibitem[{Liu et~al.(2024{\natexlab{b}})Liu, Wang, Flanigan, and Liu}]{liu2024large}
Chris~Yuhao Liu, Yaxuan Wang, Jeffrey Flanigan, and Yang Liu. 2024{\natexlab{b}}.
\newblock Large language model unlearning via embedding-corrupted prompts.
\newblock \emph{arXiv preprint arXiv:2406.07933}.

\bibitem[{Liu et~al.(2024{\natexlab{c}})Liu, Yao, Jia, Casper, Baracaldo, Hase, Xu, Yao, Li, Varshney et~al.}]{liu2024rethinking}
Sijia Liu, Yuanshun Yao, Jinghan Jia, Stephen Casper, Nathalie Baracaldo, Peter Hase, Xiaojun Xu, Yuguang Yao, Hang Li, Kush~R Varshney, et~al. 2024{\natexlab{c}}.
\newblock Rethinking machine unlearning for large language models.
\newblock \emph{arXiv preprint arXiv:2402.08787}.

\bibitem[{Liu et~al.(2024{\natexlab{d}})Liu, Dou, Tan, Tian, and Jiang}]{liu2024towards}
Zheyuan Liu, Guangyao Dou, Zhaoxuan Tan, Yijun Tian, and Meng Jiang. 2024{\natexlab{d}}.
\newblock Towards safer large language models through machine unlearning.
\newblock \emph{arXiv preprint arXiv:2402.10058}.

\bibitem[{Lu et~al.(2022)Lu, Welleck, Hessel, Jiang, Qin, West, Ammanabrolu, and Choi}]{lu2022quark}
Ximing Lu, Sean Welleck, Jack Hessel, Liwei Jiang, Lianhui Qin, Peter West, Prithviraj Ammanabrolu, and Yejin Choi. 2022.
\newblock Quark: Controllable text generation with reinforced unlearning.
\newblock \emph{Advances in neural information processing systems}, 35:27591--27609.

\bibitem[{Maini et~al.(2024)Maini, Feng, Schwarzschild, Lipton, and Kolter}]{maini2024tofu}
Pratyush Maini, Zhili Feng, Avi Schwarzschild, Zachary~C. Lipton, and J.~Zico Kolter. 2024.
\newblock \href {https://arxiv.org/abs/2401.06121} {Tofu: A task of fictitious unlearning for llms}.
\newblock \emph{Preprint}, arXiv:2401.06121.

\bibitem[{Meng et~al.(2022)Meng, Bau, Andonian, and Belinkov}]{meng2022locating}
Kevin Meng, David Bau, Alex Andonian, and Yonatan Belinkov. 2022.
\newblock Locating and editing factual associations in gpt.
\newblock \emph{Advances in Neural Information Processing Systems}, 35:17359--17372.

\bibitem[{Motoki et~al.(2023)Motoki, Pinho~Neto, and Rodrigues}]{motoki2023more}
Fabio Motoki, Valdemar Pinho~Neto, and Victor Rodrigues. 2023.
\newblock More human than human: Measuring chatgpt political bias.
\newblock \emph{Available at SSRN 4372349}.

\bibitem[{Pawelczyk et~al.(2023)Pawelczyk, Neel, and Lakkaraju}]{pawelczyk2023context}
Martin Pawelczyk, Seth Neel, and Himabindu Lakkaraju. 2023.
\newblock In-context unlearning: Language models as few shot unlearners.
\newblock \emph{arXiv preprint arXiv:2310.07579}.

\bibitem[{Puigcerver et~al.(2022)Puigcerver, Jenatton, Riquelme, Awasthi, and Bhojanapalli}]{puigcerver2022adversarial}
Joan Puigcerver, Rodolphe Jenatton, Carlos Riquelme, Pranjal Awasthi, and Srinadh Bhojanapalli. 2022.
\newblock On the adversarial robustness of mixture of experts.
\newblock \emph{arXiv preprint arXiv:2210.10253}.

\bibitem[{Riquelme et~al.(2021)Riquelme, Puigcerver, Mustafa, Neumann, Jenatton, Susano~Pinto, Keysers, and Houlsby}]{riquelme2021scaling}
Carlos Riquelme, Joan Puigcerver, Basil Mustafa, Maxim Neumann, Rodolphe Jenatton, Andr{\'e} Susano~Pinto, Daniel Keysers, and Neil Houlsby. 2021.
\newblock Scaling vision with sparse mixture of experts.
\newblock \emph{Advances in Neural Information Processing Systems}, 34:8583--8595.

\bibitem[{Shazeer et~al.(2017)Shazeer, Mirhoseini, Maziarz, Davis, Le, Hinton, and Dean}]{shazeer2017outrageously}
Noam Shazeer, Azalia Mirhoseini, Krzysztof Maziarz, Andy Davis, Quoc Le, Geoffrey Hinton, and Jeff Dean. 2017.
\newblock Outrageously large neural networks: The sparsely-gated mixture-of-experts layer.
\newblock \emph{arXiv preprint arXiv:1701.06538}.

\bibitem[{Shen et~al.(2023)Shen, Zhang, Cao, Tan, Chen, and Gan}]{shen2023moduleformer}
Yikang Shen, Zheyu Zhang, Tianyou Cao, Shawn Tan, Zhenfang Chen, and Chuang Gan. 2023.
\newblock Moduleformer: Learning modular large language models from uncurated data.
\newblock \emph{arXiv preprint arXiv:2306.04640}.

\bibitem[{Shi et~al.(2024)Shi, Lee, Huang, Malladi, Zhao, Holtzman, Liu, Zettlemoyer, Smith, and Zhang}]{shi2024muse}
Weijia Shi, Jaechan Lee, Yangsibo Huang, Sadhika Malladi, Jieyu Zhao, Ari Holtzman, Daogao Liu, Luke Zettlemoyer, Noah~A Smith, and Chiyuan Zhang. 2024.
\newblock Muse: Machine unlearning six-way evaluation for language models.
\newblock \emph{arXiv preprint arXiv:2407.06460}.

\bibitem[{Sun et~al.(2024)Sun, Huang, Wang, Wu, Zhang, Gao, Huang, Lyu, Zhang, Li et~al.}]{sun2024trustllm}
Lichao Sun, Yue Huang, Haoran Wang, Siyuan Wu, Qihui Zhang, Chujie Gao, Yixin Huang, Wenhan Lyu, Yixuan Zhang, Xiner Li, et~al. 2024.
\newblock Trustllm: Trustworthiness in large language models.
\newblock \emph{arXiv preprint arXiv:2401.05561}.

\bibitem[{Team(2024)}]{qwen_moe}
Qwen Team. 2024.
\newblock \href {https://qwenlm.github.io/blog/qwen-moe/} {Qwen1.5-moe: Matching 7b model performance with 1/3 activated parameters"}.

\bibitem[{Thaker et~al.(2024)Thaker, Maurya, and Smith}]{thaker2024guardrail}
Pratiksha Thaker, Yash Maurya, and Virginia Smith. 2024.
\newblock Guardrail baselines for unlearning in llms.
\newblock \emph{arXiv preprint arXiv:2403.03329}.

\bibitem[{Touvron et~al.(2023)Touvron, Martin, Stone, Albert, Almahairi, Babaei, Bashlykov, Batra, Bhargava, Bhosale et~al.}]{touvron2023llama}
Hugo Touvron, Louis Martin, Kevin Stone, Peter Albert, Amjad Almahairi, Yasmine Babaei, Nikolay Bashlykov, Soumya Batra, Prajjwal Bhargava, Shruti Bhosale, et~al. 2023.
\newblock Llama 2: Open foundation and fine-tuned chat models.
\newblock \emph{arXiv preprint arXiv:2307.09288}.

\bibitem[{Wang et~al.(2024{\natexlab{a}})Wang, Wu, He, Chen, and McAuley}]{wang2024large}
Yu~Wang, Ruihan Wu, Zexue He, Xiusi Chen, and Julian McAuley. 2024{\natexlab{a}}.
\newblock Large scale knowledge washing.
\newblock \emph{arXiv preprint arXiv:2405.16720}.

\bibitem[{Wang et~al.(2024{\natexlab{b}})Wang, Chen, Dai, Xu, Li, and Wu}]{wang2024let}
Zihan Wang, Deli Chen, Damai Dai, Runxin Xu, Zhuoshu Li, and Y~Wu. 2024{\natexlab{b}}.
\newblock Let the expert stick to his last: Expert-specialized fine-tuning for sparse architectural large language models.
\newblock \emph{arXiv preprint arXiv:2407.01906}.

\bibitem[{Wei et~al.(2024)Wei, Huang, Huang, Xie, Qi, Xia, Mittal, Wang, and Henderson}]{wei2024assessing}
Boyi Wei, Kaixuan Huang, Yangsibo Huang, Tinghao Xie, Xiangyu Qi, Mengzhou Xia, Prateek Mittal, Mengdi Wang, and Peter Henderson. 2024.
\newblock Assessing the brittleness of safety alignment via pruning and low-rank modifications.
\newblock \emph{arXiv preprint arXiv:2402.05162}.

\bibitem[{Wen et~al.(2023)Wen, Ke, Sun, Zhang, Li, Bai, and Huang}]{wen2023unveiling}
Jiaxin Wen, Pei Ke, Hao Sun, Zhexin Zhang, Chengfei Li, Jinfeng Bai, and Minlie Huang. 2023.
\newblock Unveiling the implicit toxicity in large language models.
\newblock In \emph{The 2023 Conference on Empirical Methods in Natural Language Processing}.

\bibitem[{Wu et~al.(2023)Wu, Li, Xu, Dong, Wu, Bian, and Xiong}]{wu2023depn}
Xinwei Wu, Junzhuo Li, Minghui Xu, Weilong Dong, Shuangzhi Wu, Chao Bian, and Deyi Xiong. 2023.
\newblock Depn: Detecting and editing privacy neurons in pretrained language models.
\newblock \emph{arXiv preprint arXiv:2310.20138}.

\bibitem[{xAI(2024)}]{grok1}
xAI. 2024.
\newblock Grok-1: Python library for interpretable machine learning with grok.
\newblock \url{https://github.com/xai-org/grok-1}.
\newblock Accessed: 2024-09-25.

\bibitem[{Yang et~al.(2024)Yang, Yang, Hui, Zheng, Yu, Zhou, Li, Li, Liu, Huang et~al.}]{yang2024qwen2}
An~Yang, Baosong Yang, Binyuan Hui, Bo~Zheng, Bowen Yu, Chang Zhou, Chengpeng Li, Chengyuan Li, Dayiheng Liu, Fei Huang, et~al. 2024.
\newblock Qwen2 technical report.
\newblock \emph{arXiv preprint arXiv:2407.10671}.

\bibitem[{Yao et~al.(2024)Yao, Chien, Du, Niu, Wang, Cheng, and Yue}]{yao2024machine}
Jin Yao, Eli Chien, Minxin Du, Xinyao Niu, Tianhao Wang, Zezhou Cheng, and Xiang Yue. 2024.
\newblock Machine unlearning of pre-trained large language models.
\newblock \emph{arXiv preprint arXiv:2402.15159}.

\bibitem[{Yao et~al.(2023)Yao, Xu, and Liu}]{yao2023large}
Yuanshun Yao, Xiaojun Xu, and Yang Liu. 2023.
\newblock Large language model unlearning.
\newblock \emph{arXiv preprint arXiv:2310.10683}.

\bibitem[{Yu et~al.(2023)Yu, Jeoung, Kasi, Yu, and Ji}]{yu2023unlearning}
Charles Yu, Sullam Jeoung, Anish Kasi, Pengfei Yu, and Heng Ji. 2023.
\newblock Unlearning bias in language models by partitioning gradients.
\newblock In \emph{Findings of the Association for Computational Linguistics: ACL 2023}, pages 6032--6048.

\bibitem[{Zhang et~al.(2023{\natexlab{a}})Zhang, Wang, Xu, Wang, and Shi}]{zhang2023forget}
Eric Zhang, Kai Wang, Xingqian Xu, Zhangyang Wang, and Humphrey Shi. 2023{\natexlab{a}}.
\newblock Forget-me-not: Learning to forget in text-to-image diffusion models.
\newblock \emph{arXiv preprint arXiv:2303.17591}.

\bibitem[{Zhang et~al.(2024)Zhang, Lin, Bai, and Mei}]{zhang2024negative}
Ruiqi Zhang, Licong Lin, Yu~Bai, and Song Mei. 2024.
\newblock Negative preference optimization: From catastrophic collapse to effective unlearning.
\newblock \emph{arXiv preprint arXiv:2404.05868}.

\bibitem[{Zhang et~al.(2023{\natexlab{b}})Zhang, Cai, Chen, Zhang, Zhang, Chen, Chang, Wang, and Liu}]{zhang2023robust}
Yihua Zhang, Ruisi Cai, Tianlong Chen, Guanhua Zhang, Huan Zhang, Pin-Yu Chen, Shiyu Chang, Zhangyang Wang, and Sijia Liu. 2023{\natexlab{b}}.
\newblock Robust mixture-of-expert training for convolutional neural networks.
\newblock In \emph{Proceedings of the IEEE/CVF International Conference on Computer Vision}, pages 90--101.

\bibitem[{Zhang et~al.(2021)Zhang, Lin, Liu, Li, Sun, and Zhou}]{zhang2021moefication}
Zhengyan Zhang, Yankai Lin, Zhiyuan Liu, Peng Li, Maosong Sun, and Jie Zhou. 2021.
\newblock Moefication: Transformer feed-forward layers are mixtures of experts.
\newblock \emph{arXiv preprint arXiv:2110.01786}.

\bibitem[{Zhou et~al.(2022)Zhou, Lei, Liu, Du, Huang, Zhao, Dai, Le, Laudon et~al.}]{zhou2022mixture}
Yanqi Zhou, Tao Lei, Hanxiao Liu, Nan Du, Yanping Huang, Vincent Zhao, Andrew~M Dai, Quoc~V Le, James Laudon, et~al. 2022.
\newblock Mixture-of-experts with expert choice routing.
\newblock \emph{Advances in Neural Information Processing Systems}, 35:7103--7114.

\bibitem[{Zhu et~al.(2024)Zhu, Qu, Dong, Ruan, Tong, He, and Cheng}]{zhu2024llama}
Tong Zhu, Xiaoye Qu, Daize Dong, Jiacheng Ruan, Jingqi Tong, Conghui He, and Yu~Cheng. 2024.
\newblock Llama-moe: Building mixture-of-experts from llama with continual pre-training.
\newblock \emph{arXiv preprint arXiv:2406.16554}.

\bibitem[{Zoph et~al.(2022{\natexlab{a}})Zoph, Bello, Kumar, Du, Huang, Dean, Shazeer, and Fedus}]{zoph2022designing}
Barret Zoph, Irwan Bello, Sameer Kumar, Nan Du, Yanping Huang, Jeff Dean, Noam Shazeer, and William Fedus. 2022{\natexlab{a}}.
\newblock Designing effective sparse expert models.
\newblock \emph{arXiv preprint arXiv:2202.08906}.

\bibitem[{Zoph et~al.(2022{\natexlab{b}})Zoph, Bello, Kumar, Du, Huang, Dean, Shazeer, and Fedus}]{zoph2022st}
Barret Zoph, Irwan Bello, Sameer Kumar, Nan Du, Yanping Huang, Jeff Dean, Noam Shazeer, and William Fedus. 2022{\natexlab{b}}.
\newblock St-moe: Designing stable and transferable sparse expert models.
\newblock \emph{arXiv preprint arXiv:2202.08906}.

\bibitem[{Zou et~al.(2023)Zou, Wang, Kolter, and Fredrikson}]{zou2023universal}
Andy Zou, Zifan Wang, J~Zico Kolter, and Matt Fredrikson. 2023.
\newblock Universal and transferable adversarial attacks on aligned language models.
\newblock \emph{arXiv preprint arXiv:2307.15043}.

\end{thebibliography}
\end{document}